\pdfoutput=1
\PassOptionsToPackage{table,xcdraw}{xcolor}
\documentclass[11pt]{article}
\usepackage[]{ACL}
\usepackage{graphicx}
\usepackage{times}
\usepackage{latexsym}
\usepackage[T1]{fontenc}
\usepackage[utf8]{inputenc}
\usepackage{microtype}
\usepackage{inconsolata}
\usepackage{amsmath}
\usepackage{amssymb}
\usepackage{booktabs}
\usepackage{tabularx}
\usepackage{placeins}
\usepackage{stfloats}
\usepackage{multirow}
\usepackage{pifont}
\usepackage{hyperref}
\usepackage{mathrsfs}
\usepackage{arydshln}
\usepackage{dashrule}
\usepackage{titlesec}
\usepackage{comment}
\usepackage{makecell}
\usepackage[ruled,vlined]{algorithm2e}
\usepackage[most]{tcolorbox}
\usepackage{colortbl}
\usepackage{subcaption}
\usepackage[normalem]{ulem}
\usepackage{tablefootnote}

\definecolor{myBlue}{HTML}{27408b}
\definecolor{myOrange}{HTML}{ba4a00}
\definecolor{myGreen}{HTML}{28b463}

\begin{document}

\setlength{\abovecaptionskip}{5pt}

\title{\textsc{TracSum}: A New Benchmark for Aspect-Based Summarization with Sentence-Level Traceability in Medical Domain}

\author{
\textbf{Bohao Chu}\textsuperscript{1} \quad
\textbf{Meijie Li}\textsuperscript{1,2,3} \quad
\textbf{Sameh Frihat}\textsuperscript{1} \quad
\textbf{Chengyu Gu}\textsuperscript{1} \\
\textbf{Georg Lodde}\textsuperscript{3} \quad
\textbf{Elisabeth Livingstone}\textsuperscript{3} \quad
\textbf{Norbert Fuhr}\textsuperscript{1} \\
\textsuperscript{1}University of Duisburg-Essen,
\textsuperscript{2}Institute for AI in Medicine (IKIM) \\
\textsuperscript{3}University Hospital Essen \\
\texttt{bohao.chu@uni-due.de}
}

\maketitle

\begin{abstract}
   While document summarization with LLMs has enhanced access to textual information, concerns about the factual accuracy of these summaries persist, especially in the medical domain. Tracing evidence from which summaries are derived enables users to assess their accuracy, thereby alleviating this concern. In this paper, we introduce \textsc{TracSum}, a novel benchmark for traceable, aspect-based summarization, in which generated summaries are paired with sentence-level citations, enabling users to trace back to the original context. First, we annotate 500 medical abstracts\footnote{We focus on abstracts because they are always publicly accessible and typically include the key medical aspects.} for seven key medical aspects, yielding 3.5K summary-citations pairs. We then propose a fine-grained evaluation framework for this new task, designed to assess the completeness and consistency of generated content using four metrics. Finally, we introduce a summarization pipeline, \textsc{Track-Then-Sum}, which serves as a baseline method for comparison. In experiments, we evaluate both this baseline and a set of LLMs on \textsc{TracSum}, and conduct a human evaluation to assess the evaluation results. The findings demonstrate that \textsc{TracSum} can serve as an effective benchmark for traceable, aspect-based summarization tasks. We also observe that explicitly performing sentence-level tracking prior to summarization enhances generation accuracy, while incorporating the full context further improves completeness. Source code and guidelines are available at \url{https://github.com/chubohao/TracSum}. 
\end{abstract}

\section{Introduction}
    \begin{figure}[t]
        \includegraphics[width=\linewidth]{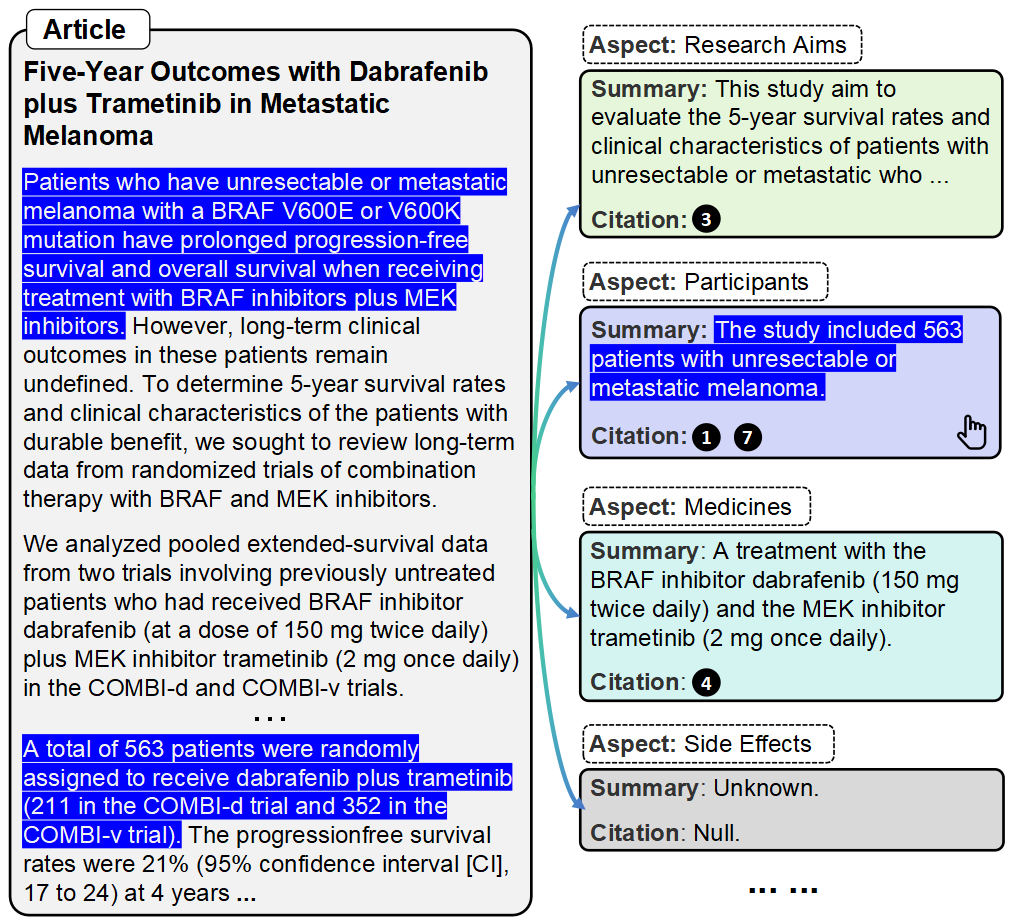}
        \caption{Schematic diagram of the \textsc{TracSum} task, where aspect-based summaries are enriched with sentence-level citations linking back to their corresponding source sentences in the medical article.}
        \setlength{\abovecaptionskip}{-10pt}
        \label{fig:medsum_overview}
        \vspace{-15pt}
    \end{figure}

    New findings observed in clinical trials are published in journal articles, which describe their design and outcomes \citep{hariton2018randomised}, serving as a crucial foundation for evidence-based medicine (EBM) \citep{sackett1997evidence, joseph-etal-2024-factpico}. Ideally, medical professionals would stay current on all medical evidence from these articles to support their decision-making, but this is impractical due to the volume and growth of the evidence base \citep{marshall2021state, frihat2024supporting}.

    
    Document summarization condenses the input document into a concise and coherent text that retains salient information \citep{narayan2018don, zheng2020two, wang2022salience, zhang-etal-2023-macsum}. Recent advancements in document summarization methods have shown promising results in generating overall summaries \citep{rush-etal-2015-neural, cheng-lapata-2016-neural, see-etal-2017-get, paulus2018deep}. However, when users refer to the same article, their areas of focus can vary significantly \citep{zhong-etal-2021-qmsum, goyal-etal-2022-hydrasum, zhang-etal-2023-macsum}. Rather than an overall summary, they are often more interested in obtaining summaries focused on specific aspects \cite{yang-etal-2023-oasum, takeshita-etal-2024-aclsum, guo-vosoughi-2024-disordered}. Therefore, generating aspect-based summaries to meet diverse user preferences is a natural and important capability for modern summarization systems \cite{xu-etal-2023-pre-trained, kolagar-zarcone-2024-humsum, takeshita-etal-2024-aclsum}. 

    Moreover, most current studies in this field \cite{zhang-etal-2023-famesumm, zhang-etal-2023-macsum, takeshita-etal-2024-aclsum} focus on unidirectional summarization with LLMs (i.e., \textit{article $\Rightarrow$ summary}). Despite their potential, state-of-the-art LLMs still struggle with factual inaccuracies \citep{mallen-etal-2023-trust, min-etal-2023-factscore}, which pose significant risks when healthcare professionals rely on these summaries for treatment decisions \citep{burns2011levels, xie-etal-2024-doclens}. By providing referenced source texts from which summaries are derived (i.e., \textit{article $\Leftarrow$ summary}), users can more easily locate relevant context and verify the generated content, thereby mitigating such concerns \citep{kambhamettu2024traceable, xie-etal-2024-doclens, deng-etal-2024-document}. Therefore, traceable summarization (i.e., \textit{article $\Leftrightarrow$ summary}) becomes especially crucial given that summarization systems can generate hallucinated content \citep{dhuliawala-etal-2024-chain}.

   To address these two concerns, we introduce \textsc{TracSum}, a novel summarization task that generates structured summaries of clinical articles across seven key medical aspects, as shown in \autoref{fig:medsum_overview}. These structured summaries not only provide flexibility to meet diverse informational needs but also enable cross-study comparisons, supporting a more comprehensive synthesis of evidence for clinical decision-making. In addition, \textsc{TracSum} extends the task by identifying the sentences cited by the summary. In real-world scenarios, this sentence-level traceable summarization enables users to locate the relevant context and verify the generation. Overall, our key contributions are as follows:  

   \vspace{0.15cm}

    \noindent\textbf{Contribution 1:} We propose \textsc{TracSum}, a novel benchmark for generating structured summaries of clinical articles across seven key aspects, enriched with sentence-level citations for each summary. To support this task, we construct a new dataset by annotating 500 clinical abstracts, resulting in 3.5K summary–citations pairs (\S\ref{sec:benchmark}). 

    \vspace{0.15cm}

    \noindent\textbf{Contribution 2:} We introduce a fine-grained automatic evaluation framework tailored for this task, which assesses the completeness and consistency of the system output by measuring the recall and precision of both generated facts and their corresponding sentence-level citations (\S\ref{sec:evaluation}). 

    \vspace{0.15cm}

    \noindent\textbf{Contribution 3:} Inspired by Chain-of-thought (CoT) reasoning \cite{wei2022chain}, we propose a summarization pipeline, \textsc{Track-Then-Sum}, which consists of a tracker $\mathcal{T}$ and a summarizer $\mathcal{S}$. The tracker $\mathcal{T}$ identifies source sentences relevant to a specific aspect, and the summarizer $\mathcal{S}$ condenses them into a short summary (\S\ref{sec:baseline}). 

    \vspace{0.15cm}

    \noindent\textbf{Contribution 4:} We evaluate a diverse set of closed- and open-source LLMs on \textsc{TracSum}, and conduct a human evaluation to assess the outputs produced by our fine-grained evaluation method. The findings demonstrate that \textsc{TracSum} can serve as an effective benchmark for traceable, aspect-based summarization in the medical domain (\S\ref{sec:experiment}).

\section{Related Work}

\subsection{Aspect-Based Summarization} \label{sec:abs}
Articles describing clinical trials often present information aligned with fixed core aspects, such as PICO\footnote{PICO: Participants/Problem (P), Intervention (I), Comparison (C), and Outcome (O).} elements \citep{richardson1995well, schardt2007utilization, schiavenato2021pico}, which represent essential components of medical evidence \citep{jin-szolovits-2018-pico, joseph-etal-2024-factpico}. Generating structured summaries for these elements offers flexibility to address diverse informational needs and facilitates cross-study comparisons \cite{yang-etal-2023-oasum, takeshita-etal-2024-aclsum}, enabling a comprehensive synthesis of evidence for clinical decision-making. To support fine-grained summarization, this work focuses on generating structured summaries that cover seven medical aspects commonly reported in clinical articles.

\subsection{Traceable Summarization} 
Identifying the citations that summaries rely on can help users verify their accuracy \cite{gao-etal-2023-enabling, xie-etal-2024-doclens}, particularly in high-stakes domains such as medicine. To support critical examination of summaries and their underlying sources, \citet{kambhamettu2024traceable} introduced a simple interaction primitive called ``traceable text.'' In the domain of Question Answering (QA), \citet{gao-etal-2023-enabling} showed that enabling LLMs to generate text with passage-level citations improves factual correctness and verifiability. Moreover, several studies on retrieval-augmented generation (RAG) approaches can support document- or paragraph-level traceability \citep{wang-etal-2024-rag, xu-etal-2024-unsupervised, wang-etal-2024-retrieval}. Building on this prior work, our research introduces sentence-level traceability of summaries generated by summarization systems, allowing users to directly inspect the source content that supports each summarized aspect.

\section{\textsc{TracSum} Benchmark} \label{sec:benchmark}

    \subsection{Task Description} \label{sec:task}
    Given a clinical article and a specific medical aspect, \textsc{TracSum} requires summarization systems to generate an aspect-based summary along with the corresponding sentence-level citations from which the summary is derived. Formally, let the input article $d = [c_1, c_2, ..., c_n]$ be a sequence of uniquely indexed sentences, and let $a$ be a target aspect selected from predefined aspects $\mathcal{A}$ (\S\ref{sec:aspect-pool}). The system $\mathcal{M}({\mathcal{C}'}, sum' \, \vert \, {d,a})$ is expected to generate an aspect-specific summary $sum'$ and a set of cited sentences $\mathcal{C}' = [c'_{1}, c'_{2}, ..., c'_{k}]$, where $c'_i$ refers to the index of a sentence in $d$ that supports the summary. If the article contains no information relevant to the given aspect, the system should output $sum' \leftarrow \text{``Unknown''}$ and $\mathcal{C}' \leftarrow \text{``Null''}$.

    \subsection{Dataset Collection} \label{sec:dataset}
    \subsubsection{Medical Aspects} \label{sec:aspect-pool}
    Building on the PICO framework (\S\ref{sec:abs}), we define $\mathcal{A}$ as a set of seven medical aspects commonly reported in clinical articles (as listed in \autoref{tab:aspects}).

    \begin{table}[h]
        \small
        \centering
        \begin{tabularx}{0.49\textwidth}{c c c}
        \toprule
        \textbf{Symbol} & \textbf{Aspect} & \textbf{Description} \\
        \midrule
        A & Aims & Objective \\
        \midrule
        I & Intervention & Treatment Method \\
        \midrule
        O & Outcomes & Results of Predefined Variables \\
        \midrule
        P & Participants & E.g., Diseases, Number \\
        \midrule
        M & Medicine & E.g., Name, Dosage \\
        \midrule
        D & Duration & Treatment Duration \\
        \midrule
        S & Side Effects & Observed Adverse Events \\
     
        \bottomrule
        \end{tabularx}
        \caption{Definition of seven medical aspects.}
        \label{tab:aspects}
        \vspace{-15pt}
    \end{table}


    
    \subsubsection{Source Articles} 
    We initially screened 741 medical abstracts from PubMed\footnote{\url{https://pubmed.ncbi.nlm.nih.gov/}}, of which 500 were ultimately included. The screening criteria were as follows: (1) the study focuses on melanoma; (2) the publication date is within the past 10 years; (3) the article is written in English; (4) the study is classified as either a Clinical Trial or a Randomized Controlled Trial; and (5) the article is published in a journal ranked in Q1 or Q2 according to the Journal Citation Reports (JCR) \citep{JCR2024}.

\begin{figure*}[t] 
    \includegraphics[width=\linewidth]{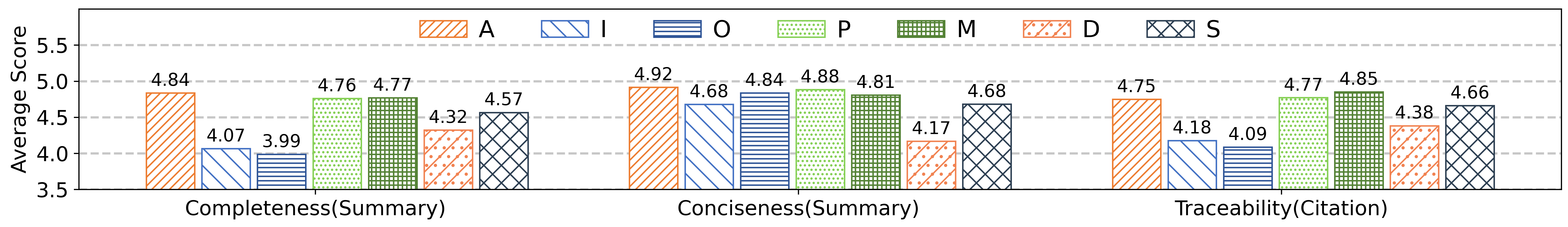} 
    \caption{Human evaluation results (5-point scale) across three qualitative metrics for the seven medical aspects. Completeness and Conciseness for summary evaluation, and Traceability for citation evaluation.}
    \setlength{\abovecaptionskip}{-10pt}
    \label{fig:human_eval} 
    \vspace{-15pt}
\end{figure*}
    \subsubsection{Initial Generation With Mistral Large}\label{sec:collecting}
    Manual dataset annotation is often costly and susceptible to stylistic inconsistencies. Consequently, leveraging LLMs to generate supervised datasets has gained popularity due to their strong zero-shot performance \cite{chen2024benchmarking, asai2024selfrag}. In this work, we automatically constructed a draft dataset by prompting Mistral Large \cite{mistral2025} to summarize 500 included abstracts, resulting in 3.5K summary–citations pairs, which were subsequently evaluated by human experts using three qualitative metrics (\S\ref{sec:phase-i}). The prompt structure comprises an abstract, a target aspect, and a type-specific instruction, followed by two demonstration examples. If the abstract lacks relevant information for the specified aspect, the model is instructed to return ``Unknown'' without generating any alternative response. An example of prompt templates is illustrated in \autoref{tab:llm_prompt} in \S\ref{sec:llm_prompt}.


    \subsubsection{Annotation Process} \label{sec: annotation_protocol}
    
    We recruited six annotators, including three medical students and three NLP researchers, who were compensated in accordance with minimum wage standards in Germany. The annotation process was carried out in two phases. In the first phase, annotators independently evaluated all data instances. In the second phase, data instances that received lower evaluation scores were manually revised. The full annotation guideline is described in \S\ref{sec:annotation_tool}. 

    \vspace{0.15cm}

    \noindent\textbf{Phase I: Evaluation.} \label{sec:phase-i}
    To ensure consistency in writing style, each data instance was independently evaluated by two independent annotators, one from the medical domain and one from the NLP domain. The annotators assessed each data instance using three qualitative evaluation metrics (as shown in \autoref{tab: metrics}) on a 5-point Likert scale, as detailed in \S\ref{sec:rating}. Evaluating a single article typically takes 10–15 minutes, depending on its complexity.

     \begin{table}[h]
        \small
        \begin{tabularx}{0.49\textwidth}{l X}
        \toprule
        \textbf{Metric} & \textbf{Description} \\ 
        \midrule
        Completeness & Does the generated summary include all facts for the given aspect? \\ 
        \midrule
        Conciseness & Does the generated summary include any irrelevant or erroneous information? \\ 
        \midrule
        Traceability & Do the citations accurately and sufficiently ground the generated summary? \\ 
        \bottomrule
        \end{tabularx}
        \caption{Qualitative evaluation metrics.}
        \label{tab: metrics}
        \vspace{-10pt}
    \end{table}

    \noindent\textbf{Phase II: Revision.}
    Out of the 3.5K evaluated data instances, we filtered out 741 (21\%) that required further revision. The filtering criteria were as follows: (1) the mean score for any of the three evaluation metrics was below 3.5, or (2) the score difference between annotators exceeded 2.0. Annotators were then instructed to revise both the summaries and their corresponding citations, as illustrated in \autoref{fig:annotation_tool_3} in \S\ref{sec:annotation_tool}.

    \subsection{Quality Analysis}
    To analyze the dataset's quality, we conducted a statistical analysis of the human evaluation results. Before filtering, the scores across all aspects and metrics are generally above 4.0 (as shown in \autoref{fig:human_eval}), indicating high overall quality. Of the 741 (21\%) filtered instances, 197 concern the O (Outcomes), 174 the I (Intervention), and 171 the D (Duration), suggesting that Mistral Large's summaries diverge most from human judgment on these three aspects, possibly due to the relatively complex information in the source texts. To assess inter-annotator agreement (IAA), we report \textit{exact match accuracy}, \textit{within-one accuracy}, and \textit{mean absolute error}, following prior work \cite{attali2006automated, zhang2007ml}. The statistical analysis revealed high agreement under the \textit{within-one accuracy} metric (84.9\%), despite a lower \textit{exact match accuracy} (66.6\%) and a \textit{mean absolute error} of 0.56, indicating acceptable consistency with only minor scoring discrepancies.

    \subsection{Characteristics of the Dataset}
    Among the 500 abstracts, the average length is 319.89 tokens, with abstract lengths ranging from 25 to 1,104 tokens. Each abstract contains an average of 10.42 sentences, spanning from 1 to 32. In the dataset of 3.5K data instances, 2,862 are positive and 638 are negative\footnote{Negative samples correspond to cases where both the summary and citation content are null.}. The positive summaries average 28.06 tokens in length, with a range from 3 to 77 tokens. On average, each positive summary cites 1.78 sentences, with a range from 1 to 7. Example data instances are presented in \autoref{tab:data_sample} (see \S\ref{sec:data_sample}), and more characteristics are described in \S\ref{sec:characteristics_dataset}.
    
    
    \begin{figure*}[t]
        \includegraphics[width=\linewidth]{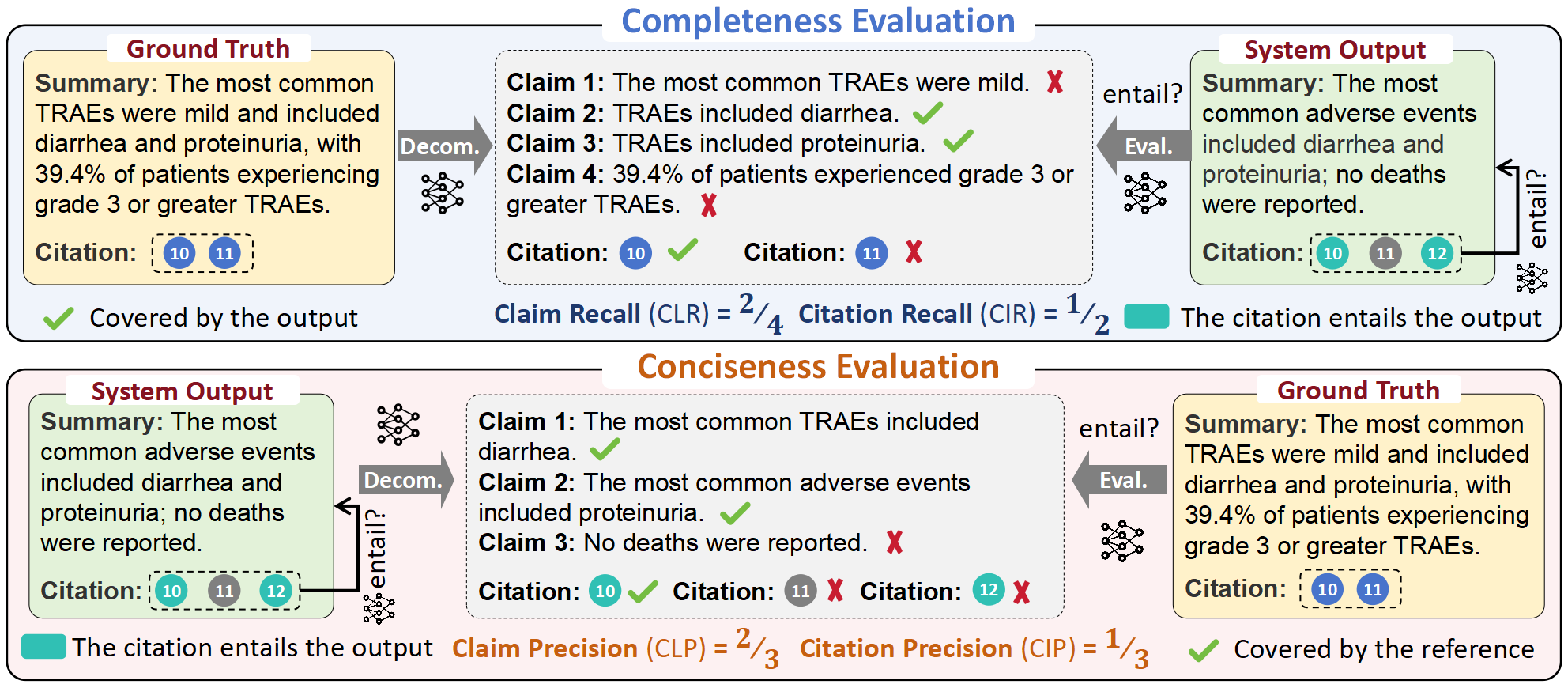}
        \caption{Overview of the automatic evaluation framework. Completeness is assessed using \textcolor{myBlue}{Claim Recall} and \textcolor{myBlue}{Citation Recall}, while conciseness is measured by \textcolor{myOrange}{Claim Precision} and \textcolor{myOrange}{Citation Precision}. Decom. denotes the claim decomposition model, and Eval. refers to the entailment evaluator.}
        \label{fig:evaluation_framework}
        \vspace{-15pt}
    \end{figure*}

\section{Automatic Evaluation Framework} \label{sec:evaluation}
Clinical texts have two essential characteristics: (1) \textit{it must be entirely complete, with no omissions} and (2) \textit{it must be fully accurate, without any errors} \cite{gao-etal-2023-enabling, xie-etal-2024-doclens}. In line with these considerations, we propose a fine-grained evaluation framework for this new task by extending the methodology of \citet{xie-etal-2024-doclens} and \citet{gao-etal-2023-enabling}, which evaluate completeness (\S\ref{sec:completeness}) and conciseness (\S\ref{sec:conciseness}) of generated content through a suite of metrics, as illustrated in \autoref{fig:evaluation_framework}. Unlike their original definitions, our approach incorporates citation recall and precision to evaluate completeness and conciseness. Before computing these metrics, we first check whether the cited sentences entail the generated summary.


\subsection{Completeness Evaluation} \label{sec:completeness}
Building on characteristic (1) of clinical texts, we evaluate completeness --- the extent to which clinically significant information is preserved in the system output. Unlike previous work \cite{van2023clinical}, which assigns an overall score, our approach emphasizes identifying which specific salient information is retained or omitted. As described in \S\ref{sec:task}, \textsc{TracSum} requires a summarization system to produce both a summary and its associated citations. To evaluate completeness, we introduce claim recall to assess summary content and citation recall to assess citation coverage. 

\vspace{0.15cm}

\noindent\textbf{Claim Recall:} Following \textsc{DocLens} \cite{xie-etal-2024-doclens}, we decompose each reference into a list of atomic subclaims using a decomposition model, where each subclaim represents a single factual statement from the reference. Let $y$ denote the reference, $\mathcal{L}_y$ the set of reference subclaims, and $y'$ the system-generated summary. We employ a natural language inference (NLI) model to evaluate whether each subclaim $l \in \mathcal{L}_y$ is entailed by $y'$. Claim recall is computed as $\frac{1}{|\mathcal{L}_y|} \sum_{l \in \mathcal{L}_y} \mathbb{I}[y'\Rightarrow l],$ where $\mathbb{I}[y' \Rightarrow l]$ is an indicator function that returns 1 if $y'$ entails $l$, and 0 otherwise. 

\vspace{0.15cm}

\noindent\textbf{Citation Recall:} In contrast to previous approaches \cite{gao-etal-2023-enabling, liu-etal-2023-evaluating, xie-etal-2024-doclens}, which consider citations valid if the cited sentences collectively support the summary, our method assesses whether each cited sentence independently supports the output. Let $\mathcal{C}$ be the set of citations in the reference and $\mathcal{C'}$ the set in the system output. A citation is considered recalled if it satisfies the following two conditions: (1) the cited sentence supports the generated summary ($c \rightarrow y'$); and (2) the citation is present in the reference ($c \in \mathcal{C}$). Citation recall is formally defined as $ \frac{1}{|\mathcal{C}|} \sum_{c\in\mathcal {C'}} \mathbb{I}[c\in\mathcal {C} \land c \rightarrow y'] $.


\subsection{Conciseness Evaluation} \label{sec:conciseness}
In line with characteristic (2), an ideal system output should avoid redundant or incorrect information. We evaluate conciseness as the proportion of generated content that is both factually accurate and salient. To this end, we use two metrics: claim precision, which assesses the informativeness and factual accuracy of the summary, and citation precision, which captures citation redundancy.

\vspace{0.15cm}

\noindent\textbf{Claim Precision:} Analogous to claim recall, we first decompose the generated summary into a list of subclaims, then use an evaluator to compute the proportion of these subclaims that are entailed by the reference. Claim precision is defined as $\frac{1}{|\mathcal{L^{'}}_y|} \sum_{l \in \mathcal{L^{'}}_y} \mathbb{I}[y\Rightarrow l]$, where $\mathcal{L^{'}}_y$ denotes the set of subclaims extracted from the generated summary. 

\vspace{0.15cm}

\noindent\textbf{Citation Precision:} To assess whether the output includes unnecessary citations, we introduce citation precision. In line with citation recall, a citation is deemed valid if it satisfies both previously defined conditions ($c \in \mathcal{C} \land c \rightarrow y'$). Citation precision is then calculated as the proportion of system-generated citations that fulfill these criteria.

\section{Baseline Method} \label{sec:baseline}
In this section, we introduce our baseline method, \textsc{Track-Then-Sum} (TTS), which consists of a tracker $\mathcal{T}$ and a summarizer $\mathcal{S}$ (available in two variants), as illustrated in \autoref{fig:Track-Then-Sum} in \S\ref{sec:generation_pipelines}. The training procedure is detailed in \S\ref{sec:tts}.

\subsection{Inference Overview}
The \textsc{Track-Then-Sum} generation pipeline contains two phases: tracking and summarization. In the first phase, $\mathcal{T}$ identifies the sentences most relevant to the given aspect. In the second phase, $\mathcal{S}$ generates a concise summary based on the selected sentences. Finally, the summary and citations are merged into the output, as shown in \hyperref[tab:algorithm1]{Algorithm 1}.
\begin{table}[t]
    \small
    \begin{tabularx}{0.49\textwidth}{l}
    \toprule
    \textbf{Algorithm 1:} \textsc{Track-Then-Sum} Inference \\ 
    \midrule
    \makecell[l]{\textbf{Require:} Tracker $\mathcal{T}$, Summarizer $\mathcal{S}$} \\ 
    
    \textbf{Input:} article $d=\{c_{1}, c_2, ...,c_{n}\}$ and aspect $a \in \mathcal{A}$  \\ 
    \textbf{Output:} summary $sum$ and its citations $\mathcal{C}'$ \\
    1: $\mathcal{C}' \leftarrow \emptyset$; \\
    2: \textbf{\textcolor{myBlue}{foreach}} $c \in \{c_{1}, c_2, ...,c_{n}\}$  \\
    3: \ \ \ \ \ \ $\mathcal{T}$ predict \textbf{\textcolor{myOrange}{relevance}} given $(a, c)$; \\
    4: \ \ \  \ \ \ \textbf{\textcolor{myBlue}{if}} \textbf{\textcolor{myOrange}{relevance}} == Yes \textbf{\textcolor{myBlue}{then}} append $c$ to $\mathcal{C}'$; \\
    5: summary $sum \leftarrow \mathcal{S}(a, \mathcal{C'}) \ or \ \mathcal{S}(a, (\mathcal{C'} \oplus f.))$; \\
    \bottomrule
    \end{tabularx}
    \caption*{Algorithm 1: \textsc{Track-Then-Sum} inference process. }
    \vspace{-15pt}
\end{table}\label{tab:algorithm1}

\subsection{Tracker $\mathcal{T}$}
\textbf{Data Collection:} We first applied sentence tokenization to each abstract in the training set. For each sentence, we generated \((c, a)\) pairs by combining it with every predefined aspect $a \in \mathcal{A}$. Each pair was labeled with a binary variable \(y\) based on the corresponding \textit{citations} field: if the sentence index appeared in the \textit{citations} associated with aspect $a$, we assigned \(y=1\); otherwise, \(y=0\). The resulting training dataset is denoted as $\mathcal{D}_{\mathcal{T}}$. 
\vspace{0.15cm}

\noindent\textbf{Training:} Given the constructed dataset \(\mathcal{D}_{\mathcal{T}}\), we initialized tracker \(\mathcal{T}\) using a pre-trained language model (LM) as the backbone. The model was subsequently fine-tuned on \(\mathcal{D}_{\mathcal{T}}\) using a standard binary classification objective which maximizes the log-likelihood of the observed labels: 
\begin{equation*}
\max_{\mathcal{T}} \mathbb{E}_{((c, a),y)\sim\mathcal{D}_{\mathcal{T}}} \log p_{\mathcal{T}}(y \mid (c, a))
\end{equation*}

\subsection{Summarizer $\mathcal{S}$}

\textbf{Data Collection:}  For each summary $sum$ in the training set, we extracted related sentences from the abstract based on the \textit{citations} field to form the set $\mathcal{C}$. Each $\mathcal{C}$ was paired with its associated aspect $a$, and combined with the $sum$ to form $((\mathcal{C}, a), sum)$. The resulting training dataset is denoted as $\mathcal{D}_{\mathcal{S}}$.

\vspace{0.15cm}

\noindent\textbf{Training:} Similar to the training of $\mathcal{T}$, we initialized summarizer $\mathcal{S}$ using a pre-trained LM as the backbone. We then fine-tuned summarizer $\mathcal{S}$ on $\mathcal{D}_{\mathcal{S}}$ using a standard next-token prediction objective, which maximizes the likelihood of generating the target summary $sum$ given the input $(\mathcal{C}, a)$ pair:
\begin{equation*}
    \max_{\mathcal{S}} \mathbb{E}_{((\mathcal{C}, a),sum)\sim\mathcal{D}_{\mathcal{S}}} \log p_{\mathcal{S}}(sum \mid C, a)
\end{equation*}

To investigate the impact of incorporating full context (denoted as $f.$), we trained a variant $\mathcal{S}$ that generates a summary given the input $(\mathcal{C} \oplus f., a)$.

\section{Experiment} \label{sec:experiment}
In this section, we aim to address the following research questions: \textbf{RQ1:} How effective is \textsc{TracSum} as a benchmark for evaluating LLMs in aspect-based summarization with sentence-level traceability? \textbf{RQ2:} To what extent does the proposed evaluation method align with human judgment, and what role does the evaluator play in this process? \textbf{RQ3:} Which factors most significantly impact the accuracy of traceable summarization? To address these questions, we begin by conducting a preliminary evaluation of several LLMs, including both proprietary models (e.g., GPT-4o \cite{hurst2024gpt}) and open-source models (e.g., LLaMA-3.1 \cite{grattafiori2024llama}, Mistral \cite{jiang2024mixtral}, and Gemma-3 \cite{team2025gemma}).

\subsection{Experimental Setting}
\begin{figure}[t]
    \centering
    \includegraphics[width=0.49\textwidth]{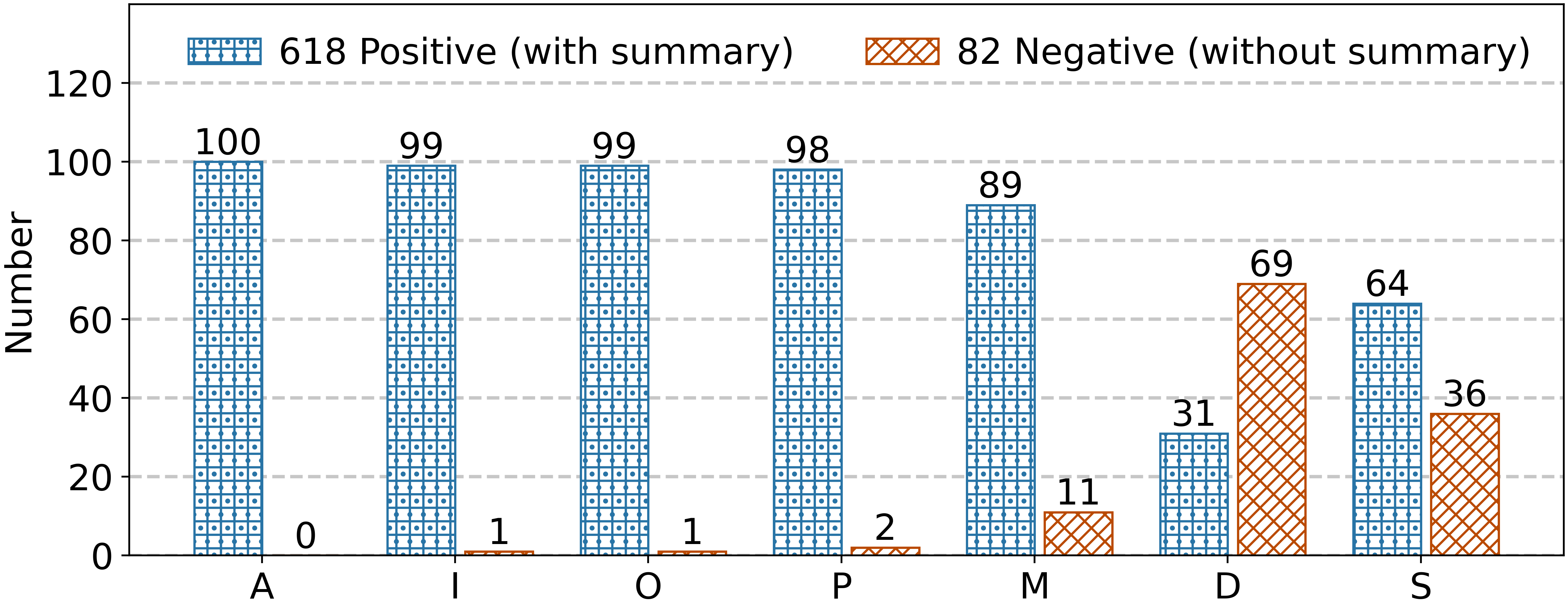}
    \caption{ Distribution of test data across seven aspects.}
    \label{fig:test-set}
    \vspace{-15pt}
\end{figure}

\noindent\textbf{Data Preparation:} The \textsc{TracSum} dataset was randomly split into training and test sets with an 8:2 ratio. We examined the distribution of samples in the test set across the seven predefined aspects, along with the proportion of positive and negative instances for each, as shown in \autoref{fig:test-set}. The results show that while nearly all abstracts contain information related to Aims (A), Intervention (I), and Outcomes (O), only 31\% explicitly mention the Duration (D) aspect. The baseline model was fine-tuned on the training set, and both the baseline and LLMs were evaluated on the test set.

\vspace{0.15cm}

\noindent\textbf{Backbone Model Selection:}
The \textsc{Track-Then-Sum} (TTS) pipeline comprises two components (Tracker $\mathcal{T}$ and Summarizer $\mathcal{S}$) that can be initialized with any pre-trained LM. For consistency and ease of deployment, we adopt Llama-3.1-8B \cite{dubey2024llama} as the backbone for both components, with the training details provided in \S\ref{sec:tts}.

\vspace{0.15cm}

\noindent\textbf{LLMs and Prompt Setting:} We selected several widely used instruction-following LLMs for evaluation, as listed in \autoref{tab:preliminary_results}. All models were evaluated using a two-shot prompting strategy, with each prompt containing one positive and one negative example. To ensure consistency, each model was prompted using its official input format with identical content (see \autoref{tab:llm_prompt} in \S\ref{sec:llm_prompt}), and a fixed temperature of 1.0 was used across all generations. Larger models were accessed via their official APIs, incurring additional usage costs (see \S\ref{sec:api_cost}).

\begin{table}[h]
    \small
    \begin{tabularx}{0.49\textwidth}{l}
    \toprule
    \textbf{Algorithm 2:} Computation Process of Evaluation Metrics \\ 
    \midrule
    \makecell[l]{\textbf{Require:} decomposition model: $\mathcal{E}$, NLI model: $\phi$} \\ 
    
    \textbf{Input:} system output $(sum', \mathcal{C}')$, reference $(sum,  \mathcal{C})$  \\ 
    \textbf{Output:} \textbf{\textcolor{myBlue}{CLR}}, \textbf{\textcolor{myBlue}{CIR}}, \textbf{\textcolor{myOrange}{CLP}}, \textbf{\textcolor{myOrange}{CIP}} \\
    1: \ \ $\{s_1, s_2, ..., s_n\} \leftarrow \mathcal{E}(sum); 0 \leftarrow n$;\\
    2: \ \ \textcolor{myBlue}{foreach} $s_i \in \{s_{1}, s_2,...,s_{n}\}$  \\
    3: \ \  \ \ \  \ \ \ \textcolor{myBlue}{if} $\phi(sum', s_i)$ == 1 \textcolor{myBlue}{then} $n$++; \\
    \rowcolor{gray!20} 4: \ \  $\textbf{\textcolor{myBlue}{CLR}} \leftarrow  n / |\{s_1, s_2, ..., s_n\}|$ \\
    
    5: \ \ $0 \leftarrow n$;\\
    6: \ \ \textcolor{myBlue}{foreach} $c'_i \in \mathcal{C}'$  \\
    7: \ \ \ \ \ \ \ \ \textcolor{myBlue}{foreach} $s'_i \in \{s'_1, s'_2, ..., s'_n\}$ \\
    8:    \ \ \ \ \ \ \ \ \ \ \ \ \ \ \ \ \textcolor{myBlue}{if} $\phi(c'_i, s'_i)$ == 1  \textcolor{myBlue}{then} $n$++; \textcolor{myBlue}{break}; \\
    \rowcolor{gray!20} 9: \ \ $\textbf{\textcolor{myBlue}{CIR}} \leftarrow n/|\mathcal{C}|; \textbf{\textcolor{myOrange}{CIP}} \leftarrow n/|\mathcal{C'}|$; \\
    
    10: \ \ $\{s'_1, s'_2, ..., s'_n\} \leftarrow \mathcal{E}(sum'); n \leftarrow 0$;\\
    11: \textcolor{myBlue}{foreach} $s'_i \in \{s'_{1}, s'_2, ...,s'_{n}\}$  \\
    12: \ \ \  \ \ \ \textcolor{myBlue}{if} $\phi(sum, s'_i)$ == 1 \textcolor{myBlue}{then} $n$++; \\
    \rowcolor{gray!20} 13: $ \textbf{\textcolor{myOrange}{CLP}} \leftarrow n/|\{s'_1, s'_2, ..., s'_n\}|$; \\
    \bottomrule
    \end{tabularx}
    \caption*{Algorithm 2: Computation process of evaluation metrics. \textcolor{myBlue}{\textbf{CLR:}} Claim Recall. \textcolor{myBlue}{\textbf{CIR:}} Citation Recall. \textcolor{myOrange}{\textbf{CLP:}} Claim Precision. \textcolor{myOrange}{\textbf{CIP:}} Citation Precision.}
    \vspace{-10pt}
\end{table}\label{tab:algorithm2}

\noindent\textbf{Evaluation Setting:} In the preliminary experiment, we adopt Mistral Large \citep{mistral2024large} as the decomposition model $\mathcal{E}$, which is used to break down both the system-generated and reference summaries into a set of atomic subclaims. For the entailment evaluation, we utilize TRUE \cite{honovich-etal-2022-true-evaluating} as the evaluator $\phi$. Let $\phi(p, h)$ denote the output of the NLI model, where the value is 1 if the premise $p$ entails the hypothesis $h$, and 0 otherwise. The computation process of the evaluation metrics is presented in \hyperref[tab:algorithm2]{Algorithm 2}.

\subsection{Preliminary Results}
\textbf{Comparison of LLMs:} \autoref{tab:preliminary_results} shows the evaluation results of various LLMs along with our proposed method (in two variants). We observe the following: (1) Larger open-source models (e.g., LLaMA-3.1-70B, Mistral-8x7B) consistently outperform smaller ones across all metrics. (2) Proprietary models like GPT-4o and GPT-4o-mini also perform well, with only small differences between them. (3) Our proposed method, fine-tuned from LLaMA-3.1-8B, shows clear improvements over both the base model and other LLMs, particularly on the two citation-based metrics CIR and CIP ($\geq74.0\%$), demonstrating their strength in identifying supporting source sentences.

\vspace{0.15cm}

\noindent\textbf{Performance on Completeness and Conciseness:} As shown in \autoref{tab:preliminary_results}, LLMs generally perform better on completeness than on conciseness, suggesting a tendency to generate content that exceeds the scope of the reference data. This may be due to full context visibility during generation, which can cause the models to include content only loosely related to the target aspects.

\vspace{0.15cm}

\noindent\textbf{Does Full Context Help?} In the TTS pipeline, we extend the input to the summarizer $\mathcal{S}$ by including not only the tracked sentences but also the full context (i.e., the abstract). This modification allows the TTS $\oplus \ f.$ variant to improve the claim recall CLR ($67.1\% 
\rightarrow 79.8\%$) of the generated summaries without substantially compromising performance on other metrics. With the tracker $\mathcal{T}$ output unchanged, the observed gains may stem from the full context offering useful explanations for abbreviations or domain-specific terminology, thereby helping $\mathcal{S}$ better interpret the tracked sentences. A detailed case analysis is provided in \S\ref{sec:full_analysis}.

\begin{table}[t]
    {\fontsize{8.8pt}{9pt}\selectfont
    \centering
    \begin{tabularx}{0.48\textwidth}{
        >{\hspace{-3pt}\centering\arraybackslash\hspace{-3pt}}l :
        >{\hspace{-3pt}\centering\arraybackslash}c
        >{\hspace{-3pt}\centering\arraybackslash}c :
        >{\hspace{-3pt}\centering\arraybackslash}c
        >{\hspace{-3pt}\centering\arraybackslash}c :
        >{\hspace{-3pt}\centering\arraybackslash}c
        >{\hspace{-3pt}\centering\arraybackslash}c
        }
        \toprule
   
        & \multicolumn{2}{c}{\textbf{\textcolor{myBlue}{{\hspace{-5pt} Completeness\hspace{-4pt}}}}} & \multicolumn{2}{c}{\textbf{\textcolor{myOrange}{{\hspace{-4pt} Conciseness}}}} & \multicolumn{2}{c}{\textbf{{\hspace{-4pt} F1 Score}}} \\
        \cmidrule(lr){2-3} \cmidrule(lr){4-5} \cmidrule(lr){6-7}
        \textbf{Method} & \textbf{CLR} & \textbf{CIR} & \textbf{CLP} & \textbf{CIP} & $F_1^{\text{cl.}}$ & $F_1^{\text{ci.}}$ \\
        \midrule
        \rowcolor{gray!20} 
        Llama-3.1-8B & 59.2 & 62.5 & 63.6 & 54.8 & 61.3 & 58.4 \\
        Llama-3.1-70B & \underline{74.7} & \underline{77.9} & \textbf{71.3} & \underline{67.7} & \underline{72.9} & \uwave{72.4} \\
        Mistral-7B & 59.1 & 59.5 & 55.5 & 48.4 & 57.4 & 53.4 \\
        Mistral-8x7B & 61.1 & 62.1 & 58.9 & 58.4 & 60.0 & 60.2 \\
        Gemma3-12B & 62.8 & 66.0 & 58.3 & 55.3 & 60.5 & 60.2 \\
        Gemma3-27B & 64.6 & 66.4 & 57.7 & 59.6 & 61.0 & 63.0 \\
        \midrule
        GPT-4o & \uwave{74.0} & \textbf{78.2} & 66.2 & 63.8 & \uwave{69.9} & 70.3 \\
        GPT-4o-mini & 67.8 & 76.0 & \uwave{67.6} & \uwave{68.4} & 67.7 & 72.0 \\
        \midrule
        \textsc{TTS} & 67.1 & \uwave{76.2} & \underline{68.4} & \textbf{77.0} & 67.8 & \textbf{76.6} \\
        \textsc{TTS} $\oplus \ f.$ & \textbf{79.8} & 74.6 & 67.2 & \underline{75.0} & \textbf{73.0} & \underline{74.8} \\

        \bottomrule
     \end{tabularx}
    \caption{Preliminary evaluation results (\%). \textbf{Bold} values indicate the best performance in each metric, \underline{underlined} values indicate the second-best, and \uwave{wave underlined} values indicate the third-best. $\oplus \ f.$ denotes the configuration where the full context is concatenated to the input of the summarizer $\mathcal{S}$. $F_1^{\text{cl.}}$ and $F_1^{\text{ci.}}$ represent the F1 scores for claim and citation prediction, respectively.}
    \label{tab:preliminary_results}
    \vspace{-10pt}
    }
\end{table}

\subsection{Agreement with Human Evaluation} \label{sec:huamn_agreement}
To address first sub-question of \textbf{RQ2}, we conducted a human evaluation and measured the agreement between human judgments and the automatic evaluation scores produced by the NLI model (TRUE) using \textit{Spearman’s correlation coefficient} ($\rho$) \cite{kendall1990rank} and \textit{Pearson’s correlation coefficient} ($r$) \cite{sheskin2003handbook}. We randomly sampled ten abstracts from the test set, and the annotator followed the procedure in \hyperref[tab:algorithm2]{Algorithm 2} to evaluate outputs from our TTS~$\oplus \ f.$, as shown in \autoref{tab:human_eval}. The results show an average Spearman’s $\rho=0.612$ and Pearson’s $r=0.577$, indicating a moderate positive correlation between automatic evaluation and human judgments. This suggests that our proposed evaluation framework aligns reasonably well with human assessments, while still leaving room for improvement. A detailed comparison of the final evaluation results is provided in \S\ref{sec:agreement_with_human}.
\begin{table}[h]
    \small
    \centering
    \begin{tabularx}{0.48\textwidth}{l}
        \toprule
        \rowcolor{gray!20} 
        \textbf{Reference:} \quad Subclaims \tikz[baseline=-0.5ex] \draw[->, thick] (0,0) -- (0.3,0) -- (0.3,-0.2); \quad Citations $\rightarrow$ 1, 5 \\
        
        1. The study included 533 patients. \\
        2. The patients were treatment-naive. \\
        3. The patients had unresectable stage III-IV melanoma.\\
        
        \midrule
        \rowcolor{gray!20} 
        \textbf{TTS $\oplus \ f.$ Output:} \quad Subclaims \tikz[baseline=-0.5ex] \draw[->, thick] (0,0) -- (0.3,0) -- (0.3,-0.2); \quad Citations $\rightarrow$ 1, 3, 5 \\
        
        1$'$. The study involved treatment-naive patients. \\ 
        2$'$. The patients had unresectable stage III-IV melanoma. \\
        3$'$. 533 patients received nivolumab plus ipilimumab. \\[0.1cm]
        
        \hdashline
        \\[-0.3cm]
        NLI : $reference \rightarrow s1', s2'$ \textcolor{green}{\textbf{\ding{51}}} $\nrightarrow s3'$ \textcolor{red}{\textbf{\ding{55}}} \quad CLR: 66.7\% \\
        Human: $reference \rightarrow s1', s2', s3'$ \quad \textcolor{green}{\textbf{\ding{51}}} \quad CLR: 100\% \\
        Reason: "533 patients" is found in the reference. \\
        \bottomrule
    \end{tabularx}
    \caption{A case comparing automatic and human evaluation of claim recall (PMID: 37307514, Aspect: P).}
    \vspace{-15pt}
    \label{tab:human_eval}
\end{table}

\subsection{Aspect-Wise Performance Analysis} 
To analyze the performance of the TTS~$\oplus \ f.$ variant across the seven aspects, we grouped the data by aspect and computed the four evaluation metrics for each group, as shown in \autoref{tab:aspect_eval}. We observed substantial variation in the model’s performance across different aspects. Notably, aspects O (Outcomes) and I (Intervention) received lower scores across all four evaluation metrics, likely because the corresponding abstracts often contain a large number of relevant sentences, making precise extraction more challenging. In contrast, aspect D (Duration) achieved relatively higher scores, possibly due to the fact that 69\% of its test instances are negative cases (i.e., both the summary and citation are null), which simplifies the task and makes correct predictions easier for the model.

\begin{table}[t]
    {\fontsize{8.8pt}{9pt}\selectfont
    \centering
    \begin{tabularx}{0.48\textwidth}{
        >{\centering\arraybackslash}c :
        >{\centering\arraybackslash}c
        >{\centering\arraybackslash}c :
        >{\centering\arraybackslash}c
        >{\centering\arraybackslash}c :
        >{\centering\arraybackslash}c
        >{\centering\arraybackslash}c
        }
        \toprule
   
        & \multicolumn{2}{c}{\textbf{\textcolor{myBlue}{Completeness}}} & \multicolumn{2}{c}{\textbf{\textcolor{myOrange}{Conciseness}}} & \multicolumn{2}{c}{\textbf{F1 Score}} \\
        \cmidrule(lr){2-3} \cmidrule(lr){4-5} \cmidrule(lr){6-7}
        \textbf{Aspect} & \textbf{CLR} & \textbf{CIR} & \textbf{CLP} & \textbf{CIP} & $F_1^{\text{cl.}}$ & $F_1^{\text{ci.}}$ \\
        \midrule

        A & \uwave{86.3} & \uwave{83.2} & 71.8 & \underline{89.8} & 78.4 & \underline{86.4} \\
        I & 69.8 & 61.4 & 51.0 & 47.6 & 58.9 & 53.4 \\
        O & 61.4 & 50.2 & 48.7 & 50.1 & 54.2 & 50.1 \\
        P & \underline{87.7} & 78.4 & \underline{80.2} & 84.2 & \underline{83.7} & 81.3 \\
        M & 85.4 & 71.9 & \uwave{75.1} & 73.3 & \uwave{79.9} & 72.6 \\
        D & \textbf{92.2} & \textbf{93.6} & \textbf{81.4} & \textbf{93.2} & \textbf{86.4} & \textbf{93.3} \\
        S & 75.8 & \underline{83.5} & 62.2 & \uwave{86.8} & 68.3 & \uwave{85.0} \\
        \midrule
        Avg. & 79.8 & 74.6 & 67.2 & 75.0 & 73.0 & 74.8 \\
        \bottomrule
    \end{tabularx}
    \caption{Aspect-wise performance of method TTS~$\oplus \ f.$}
    \label{tab:aspect_eval}
    \vspace{-15pt}
    }
\end{table}

\begin{figure*}[t]
  \centering
  \begin{subfigure}[t]{0.495\textwidth}
    \centering
    \includegraphics[width=\linewidth]{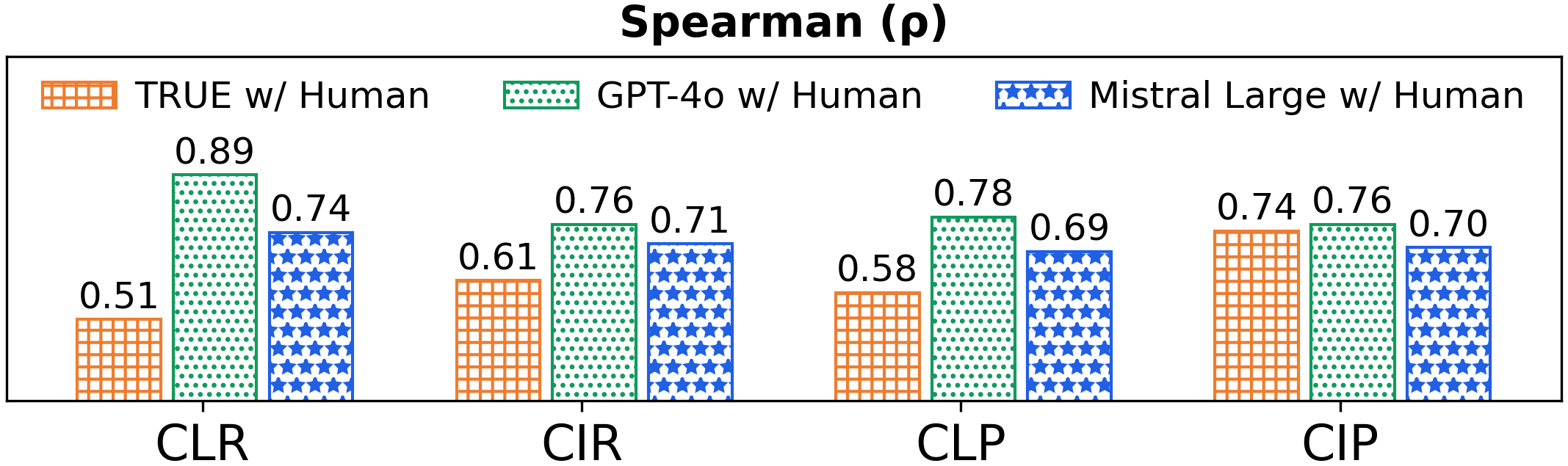}
    \vspace{-15pt}
  \end{subfigure}
  \hfill
  \begin{subfigure}[t]{0.495\textwidth}
    \centering
    \includegraphics[width=\linewidth]{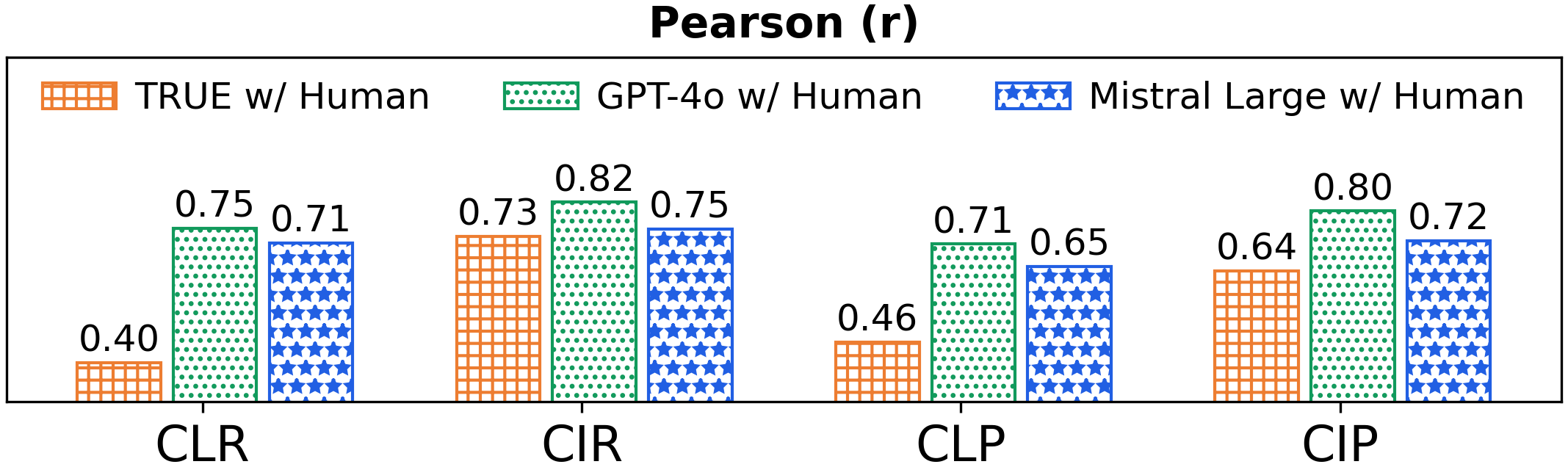}
    \vspace{-15pt}
  \end{subfigure}
   \caption{Spearman ($\rho$) and Pearson ($r$) correlations between evaluators and human scores across four metrics.}
  \label{fig:spearman}
  \vspace{-15pt}
\end{figure*} 

\subsection{Ablation Studies} 
\textbf{Comparison of Entailment Evaluators:} To address the second sub-question of \textbf{RQ2}, we experiment with two additional instruction-following LLMs as entailment evaluators: the proprietary GPT-4o \cite{hurst2024gpt} and the open-source Mistral-Large \cite{mistral2025}. Building on the experimental setup described in \S\ref{sec:huamn_agreement}, we replace the TRUE model with each of these evaluators to assess the outputs generated by the TTS~$\oplus \ f.$ variant. The experiment procedure and results are described in \S\ref{sec:comparison_evaluator}. We then compute Spearman’s $\rho$ and Pearson’s $r$ to quantify their agreement with human judgments in four metrics, as presented in \autoref{fig:spearman}. Our findings reveal that: (1) both GPT-4o ($\rho=0.80;r=0.77$) and Mistral-Large ($\rho=0.71;r=0.70$) show substantially stronger alignment with human judgments compared to TRUE ($\rho=0.61;r=0.57$); and (2) GPT-4o achieves a higher correlation with human judgments than Mistral-Large. We found that GPT-4o is better at understanding abbreviations. For instance, it correctly infers that the reference ``\textit{50 participants were randomized: 23 to observation and 27 to radiation therapy}'' entails the subclaim ``\textit{27 participants were assigned to the RT group}'', whereas Mistral and TRUE do not.

\vspace{0.15cm}

\noindent\textbf{The Effect of Tracking Order:} To address \textbf{RQ3}, we design two variants by modifying the position of the tracker $\mathcal{T}$: (i) \textsc{Sum-Then-Track (STT)} places $\mathcal{T}$ after the summarizer $\mathcal{S}$, where $\mathcal{S}$ first generates an aspect-based summary, and $\mathcal{T}$ then retrieves source sentences relevant to that summary; (ii) \textsc{End-To-End (ETE)} removes the tracker entirely and fine-tunes a single model $\mathcal{M}$ to generate both summary and citations. The experimental procedures are detailed in \S\ref{sec:generation_pipelines}. We evaluated STT and ETE on the test set, with results shown in \autoref{tab:variants_comparison}. We observe that: (1) removing the tracker results in a decline in citation-based performance, highlighting the importance of explicit sentence tracking; and (2) while STT improves claim recall, it performs worse on other metrics, likely due to its dependence on pre-generated summaries, which may introduce noise or inaccuracies. These findings emphasize the importance of incorporating tracking early in the summarization process.

\begin{table}[h]
    {\fontsize{8.8pt}{9pt}\selectfont
    \centering
    \begin{tabularx}{0.48\textwidth}{
        >{\centering\arraybackslash}l :
        >{\centering\arraybackslash}c
        >{\centering\arraybackslash}c :
        >{\centering\arraybackslash}c
        >{\centering\arraybackslash}c :
        >{\centering\arraybackslash}c
        >{\centering\arraybackslash}c
        }
        \toprule
   
        & \multicolumn{2}{c}{\textbf{\textcolor{myBlue}{{Completeness}}}} & \multicolumn{2}{c}{\textbf{\textcolor{myOrange}{{ Conciseness}}}} & \multicolumn{2}{c}{\textbf{{ F1 Score}}} \\
        \cmidrule(lr){2-3} \cmidrule(lr){4-5} \cmidrule(lr){6-7}
        \textbf{Method} & \textbf{CLR} & \textbf{CIR} & \textbf{CLP} & \textbf{CIP} & $F_1^{\text{cl.}}$ & $F_1^{\text{ci.}}$ \\
        \midrule
        TTS $\oplus \ f.$ & 79.8 & \textbf{74.6} & \textbf{67.2} & \textbf{75.0} & \textbf{73.0} & \textbf{74.8} \\
        ETE & 80.1 & 72.6 & 64.1 & 71.2 & 71.2 & 71.9 \\
        STT & \textbf{81.2} & 62.2 & 58.1 & 66.4 & 67.7 & 64.1 \\

        \bottomrule
     \end{tabularx}
    \caption{Comparison of the three tracking order variants.}
    \label{tab:variants_comparison}
    \vspace{-15pt}
    }
\end{table}

\section{Conclusion}
Motivated by growing concerns over the factual accuracy of system-generated summaries in the medical domain, we present \textsc{TracSum}, a novel benchmark for aspect-based summarization that incorporates sentence-level citations. This enables users to trace source content and verify the factual consistency of generated information. Experimental results, which show strong alignment with human judgments, demonstrate that \textsc{TracSum} can serve as a reliable benchmark for assessing both the completeness and conciseness of summaries and their citations. Furthermore, we also observe that explicitly performing sentence-level tracking prior to summarization enhances generation accuracy, while incorporating the full context further improves summary completeness.

\clearpage
\section*{Limitations}
Our research marks a significant step toward evaluating sentence-level traceability in aspect-based summarization. Nonetheless, it has certain limitations. 1). The dataset used in \textsc{TracSum} was initially generated by Mistral Large. While this approach helped reduce time and cost, it may also introduce model-specific biases. To address this concern, we implemented two mitigation strategies: (i) we conducted two rounds of human evaluation, followed by manual revision of samples with low scores or inconsistent annotations; and (ii) we excluded Mistral Large from the list of evaluated models to avoid unfair advantages or confirmation bias. 2). The structure and content of prompts can significantly influence the outputs of LLMs and, in turn, their evaluation scores. Although our prompt template was designed to be general and broadly applicable, it may not elicit the best performance from every model. To reduce potential bias and ensure fair comparison, we used a standardized prompt format across all models.

\bibliography{anthology,custom}

\newpage
\appendix

\section{Annotation Guideline} \label{sec:annotation_tool}
\subsection{Annotation Tool}
We developed a custom interactive annotation tool to support efficient and user-friendly dataset annotation, which is accessible online. The backend was implemented in the Go programming language\footnote{\url{https://go.dev/}}, chosen for its performance and simplicity. The frontend was built using the Vue.js framework\footnote{\url{https://vuejs.org/}}, which enabled a responsive and intuitive user interface, and PostgreSQL\footnote{\url{https://www.postgresql.org/}} served as the database.

\subsection{Consent Statement}
Users first register on the tool by providing their email address and selecting their role (medical domain or NLP domain). Registration is subject to approval by an administrator. During the session, only non-personal cookies are collected, and users can choose whether to accept them, as shown in \autoref{tab:consent_statment}. Access to the annotation interface is granted only after the user has provided explicit consent.

\begin{table}[h]
    \small
    \begin{tabularx}{0.49\textwidth}{l}
    \toprule
    - I agree to the use of the collected data for research purposes. \\
    - I agree to the use of functional cookies on this site.\\
    \bottomrule
    \end{tabularx}
    \caption{Consent Statement.}
    \label{tab:consent_statment}
    \vspace{-15pt}
\end{table}

\subsection{Task Assignment}
\begin{figure}[h] 
    \includegraphics[width=\linewidth]{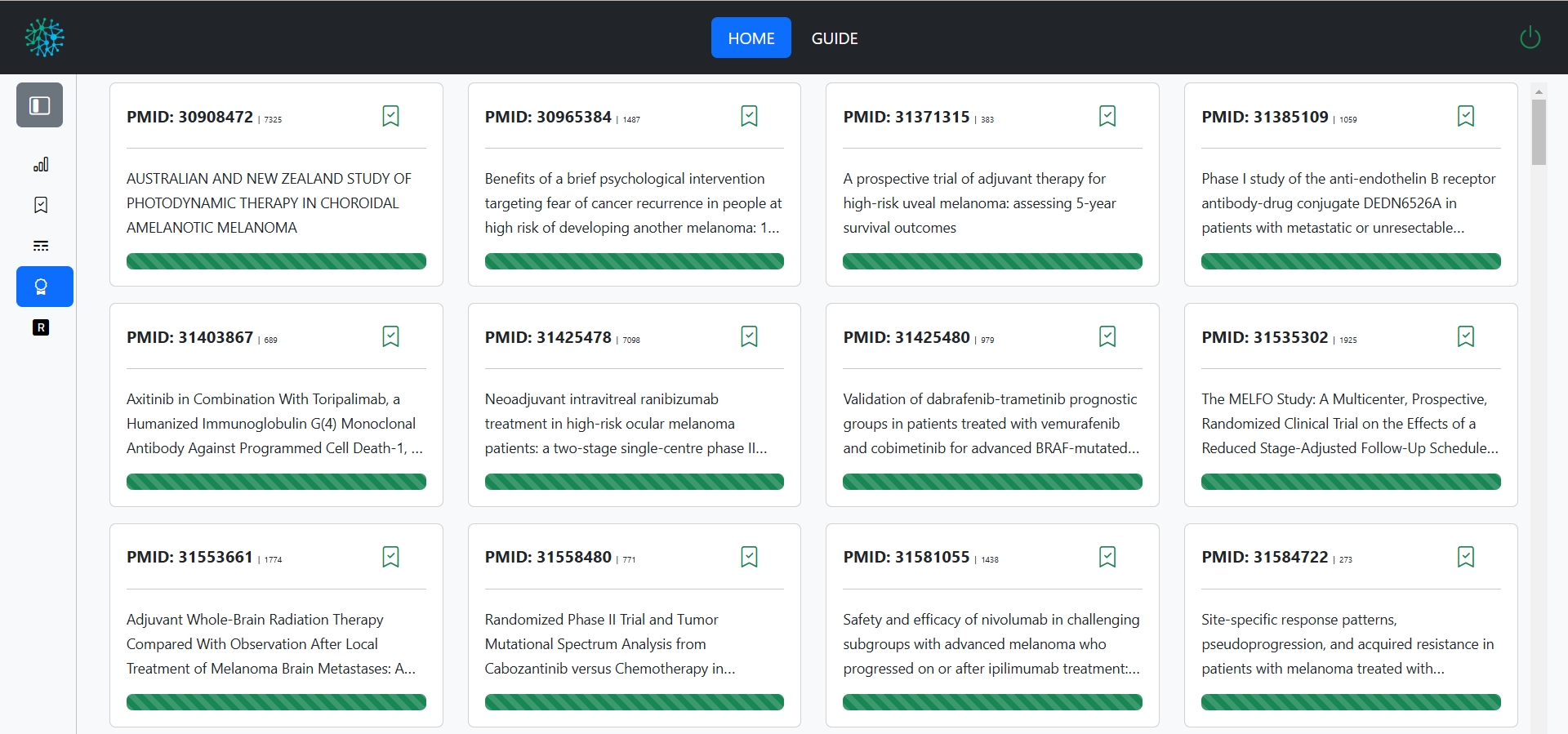} 
    \caption{List of tasks in the annotation tool.}
    \label{fig:annotation_tool_1}
    \vspace{-15pt}
\end{figure}

Both evaluation and annotation tasks are randomly assigned by administrators, as illustrated in \autoref{fig:annotation_tool_1}. Each data sample is assigned to two annotators from different domains—one from the medical domain and one from the NLP domain. Annotators were instructed not to communicate with each other to maintain data quality and ensure the authenticity of their responses.

\subsection{Evaluation Phase} \label{sec:rating}
\begin{figure}[h] 
    \includegraphics[width=\linewidth]{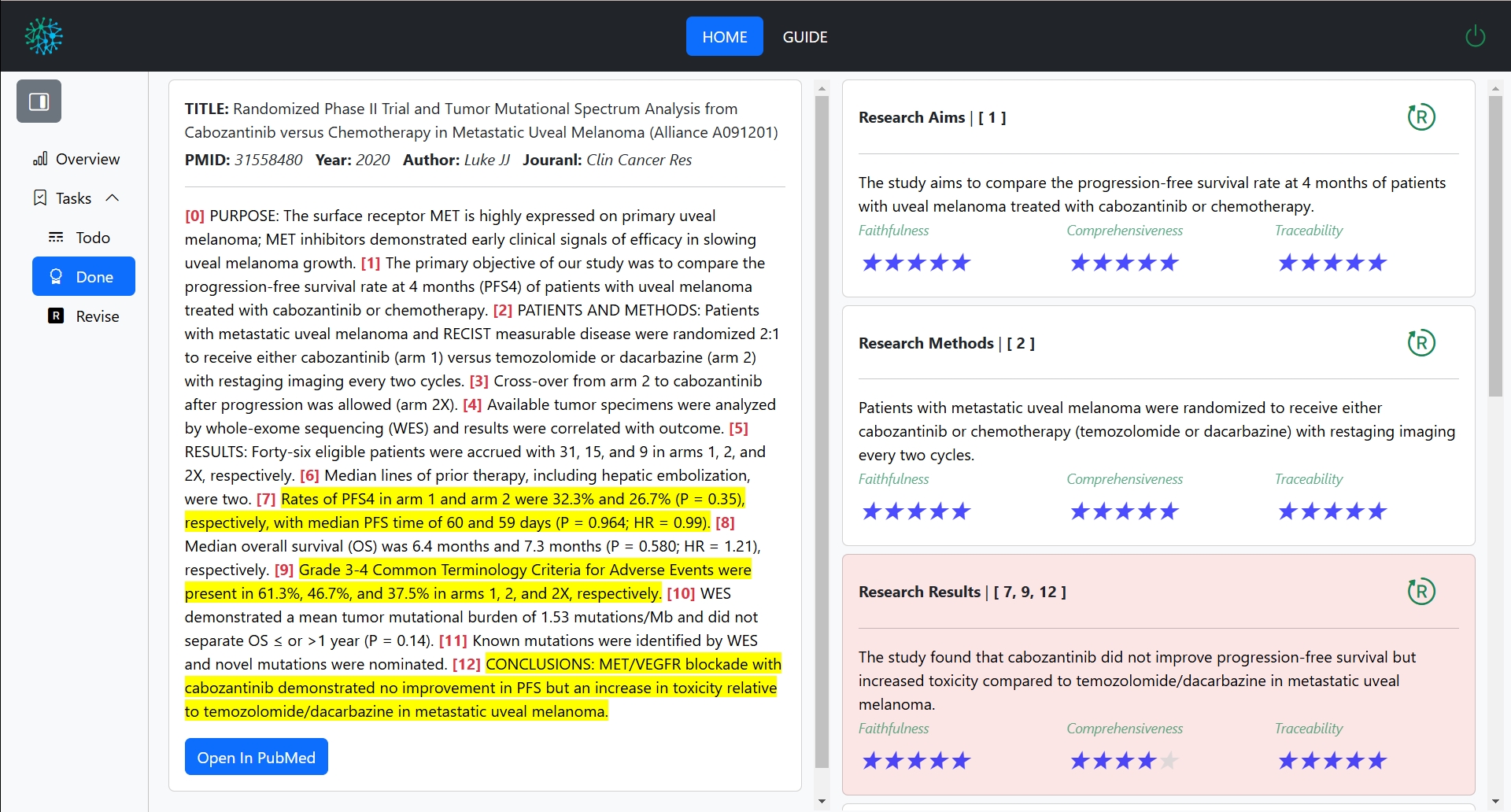} 
    \caption{Evaluation page in the annotation tool.}
    \label{fig:annotation_tool_2}
    \vspace{-15pt}
\end{figure}

n the evaluation phase, the evaluator is required to assess two components of the system output based on three aspects: Completeness (Comprehensiveness), Conciseness (Faithfulness), and Traceability. Each aspect is rated using a 5-point Likert scale, with detailed scoring guidelines provided in \autoref{tab:eval_rating}. On the evaluation page, the left panel displays the content of the article (specifically, the abstract section), while the right panel presents summary cards corresponding to seven medical aspects. When the user hovers over a summary card, the relevant sentences in the abstract on the left are highlighted, as illustrated in \autoref{fig:annotation_tool_2}. The highlight remains visible until the user hovers over another summary card, enabling easy traceability to the corresponding source sentences in the article.

\begin{table*}[t]
    \small
    \centering
    \begin{tabularx}{\textwidth}{l l l}
    \toprule
    \textbf{Aspect} & \textbf{Likert Score} & \textbf{Score Description} \\
    \midrule
    \multirow{5}{*}{Completeness}
    & \textcolor{blue}{\ding{72}\ding{72}\ding{72}\ding{72}\ding{72}} & All key relevant information from the article is accurately captured. \\
    & \textcolor{blue}{\ding{72}\ding{72}\ding{72}\ding{72}}\ding{73} & Most key relevant information from the article is present, with minor omissions. \\
    & \textcolor{blue}{\ding{72}\ding{72}\ding{72}}\ding{73}\ding{73} & Some key relevant information from the article is present, but some is missing. \\
    & \textcolor{blue}{\ding{72}\ding{72}}\ding{73}\ding{73}\ding{73} & Most key relevant information from the article is missing. \\
    & \textcolor{blue}{\ding{72}}\ding{73}\ding{73}\ding{73}\ding{73} & All key relevant information from the article is missing. \\

    \midrule
    \multirow{5}{*}{Completeness}
    & \textcolor{blue}{\ding{72}\ding{72}\ding{72}\ding{72}\ding{72}} & In the generated summary, all content is relevant to this aspect. \\
    & \textcolor{blue}{\ding{72}\ding{72}\ding{72}\ding{72}}\ding{73} & In the generated summary, most content is relevant to this aspect, with minor irrelevant parts. \\
    & \textcolor{blue}{\ding{72}\ding{72}\ding{72}}\ding{73}\ding{73} & In the generated summary, some content is relevant to this aspect, while some is irrelevant. \\
    & \textcolor{blue}{\ding{72}\ding{72}}\ding{73}\ding{73}\ding{73} & In the generated summary, most content is irrelevant to this aspect. \\
    & \textcolor{blue}{\ding{72}}\ding{73}\ding{73}\ding{73}\ding{73} & In the generated summary, all content is irrelevant to this aspect or contains errors. \\

    \midrule
    \multirow{5}{*}{Traceability}
    & \textcolor{blue}{\ding{72}\ding{72}\ding{72}\ding{72}\ding{72}} & All relevant sentences have been accurately traced (highlighted). \\
    & \textcolor{blue}{\ding{72}\ding{72}\ding{72}\ding{72}}\ding{73} & Most relevant sentences have been accurately traced (highlighted). \\
    & \textcolor{blue}{\ding{72}\ding{72}\ding{72}}\ding{73}\ding{73} & Some relevant sentences have been accurately traced, but some are missing or irrelevant. \\
    & \textcolor{blue}{\ding{72}\ding{72}}\ding{73}\ding{73}\ding{73} & Most relevant sentences have not been accurately traced. \\
    & \textcolor{blue}{\ding{72}}\ding{73}\ding{73}\ding{73}\ding{73} & None of the relevant sentences have been accurately traced. \\

    \bottomrule
    \end{tabularx}
    \caption{Evaluation Criteria and Scoring Guidelines.}
    \label{tab:eval_rating}
    \vspace{-10pt}
\end{table*}

\subsection{Revision Phase}
\begin{figure}[h] 
    \includegraphics[width=\linewidth]{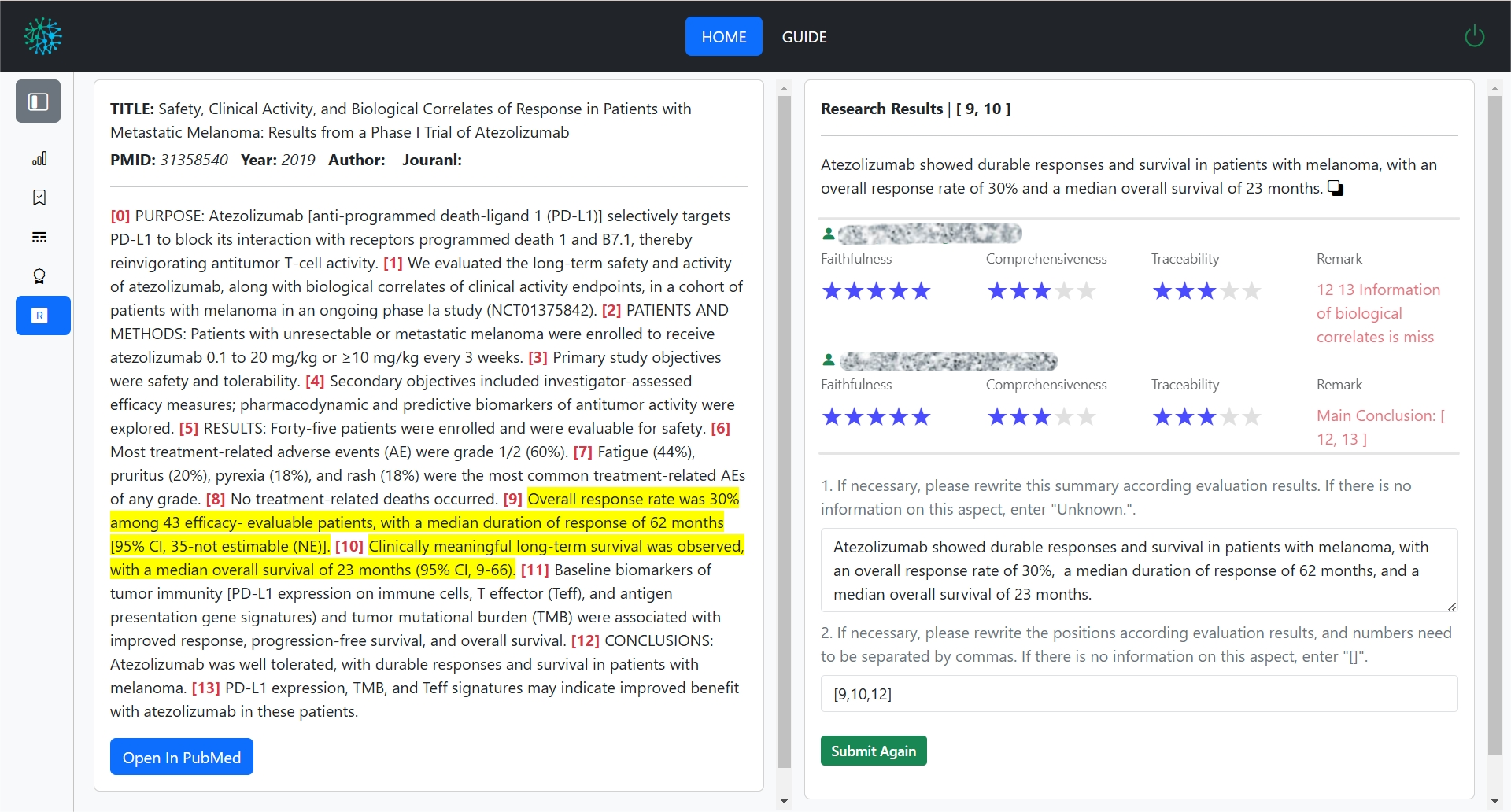} 
    \caption{Revision page in the annotation tool.}
    \label{fig:annotation_tool_3}
    \vspace{-15pt}
\end{figure}

Out of the 3.5K evaluated data instances, 741 (21\%) were filtered for further revision. The filtering criteria were as follows: (1) the mean score for any of the three evaluation metrics was below 3.5, or (2) the score difference between annotators exceeded 2.0. Annotators were then instructed to revise both the summaries and their corresponding citations based on the evaluation results. On the revision page, as illustrated in \autoref{fig:annotation_tool_3}, the left panel displayed the document content, while the right panel showed the summary along with evaluation results from two annotators. Annotators revised the summaries and updated the sentence indices according to the evaluation feedback.

\section{Characteristics of the Dataset} \label{sec:characteristics_dataset}

    \begin{figure*}[htbp]
        \centering
        \captionsetup{justification=centering}
        \begin{subfigure}[b]{0.245\textwidth}
            \includegraphics[width=\linewidth]{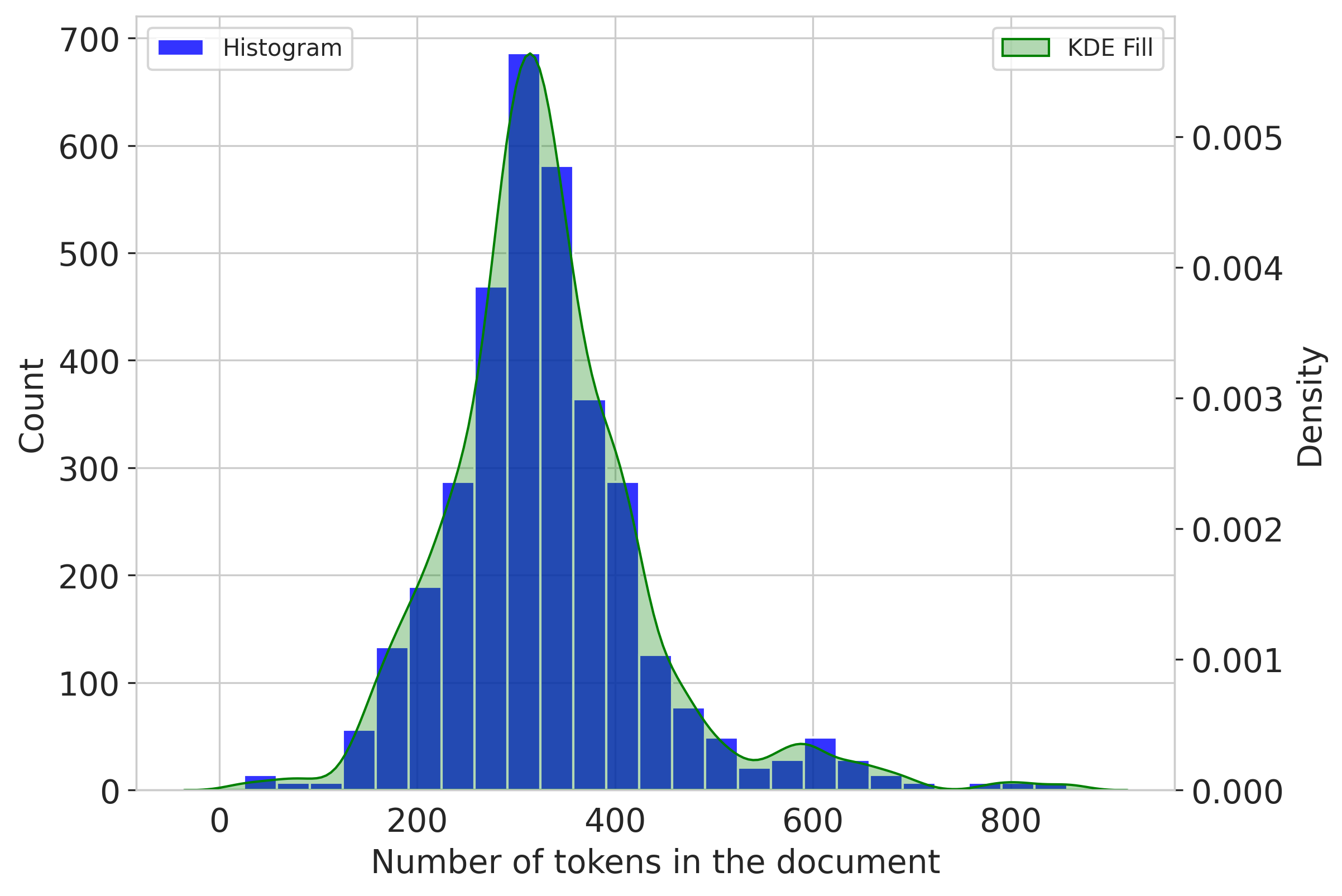}
            \caption{Distribution of token counts across abstracts.}
            \label{fig:tokens_distribution}
        \end{subfigure}
        \begin{subfigure}[b]{0.245\textwidth}
            \includegraphics[width=\linewidth]{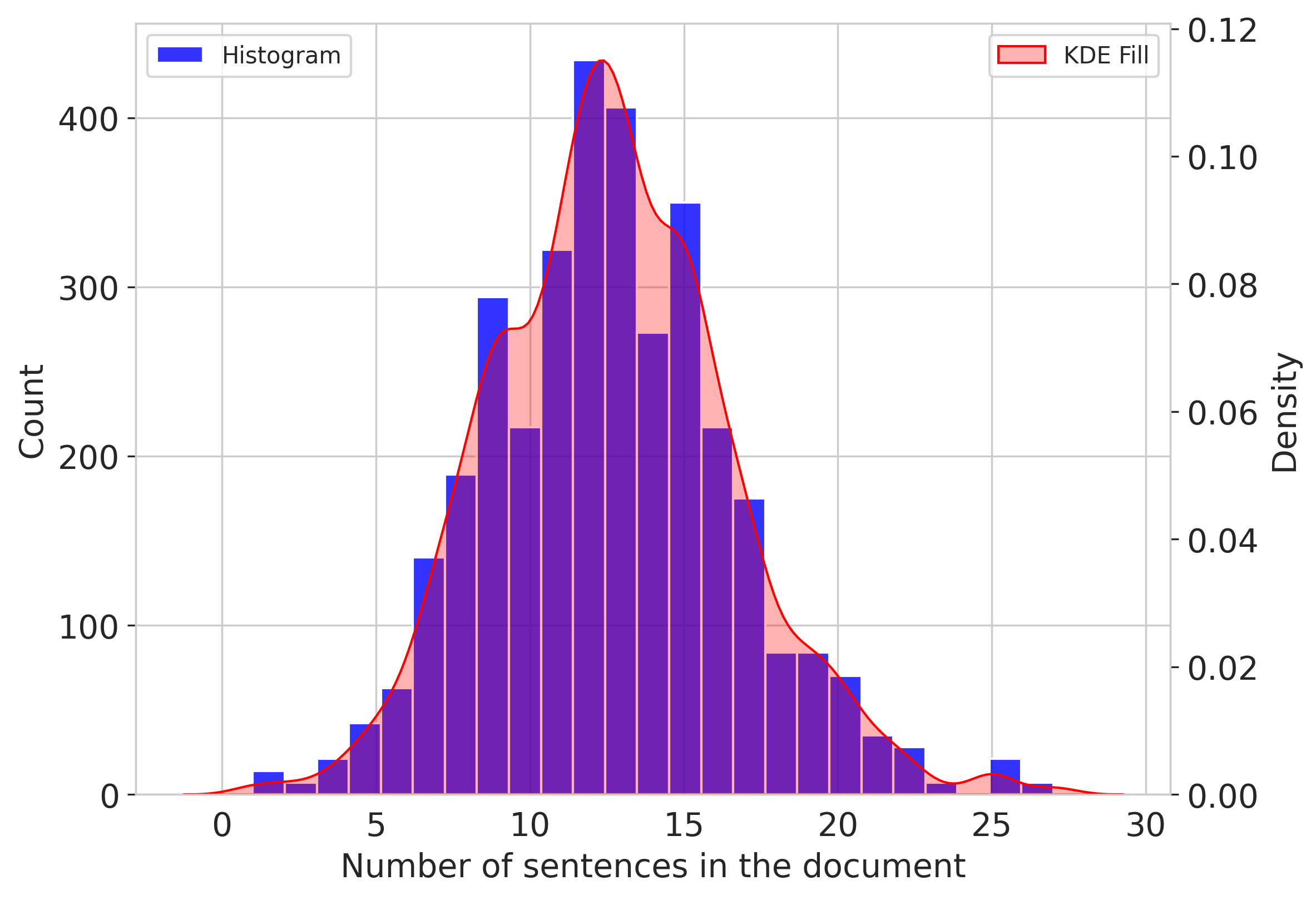}
            \caption{Distribution of sentence counts across abstracts.}
            \label{fig:sentence_distribution}
        \end{subfigure}
        \begin{subfigure}[b]{0.245\textwidth}
            \includegraphics[width=\linewidth]{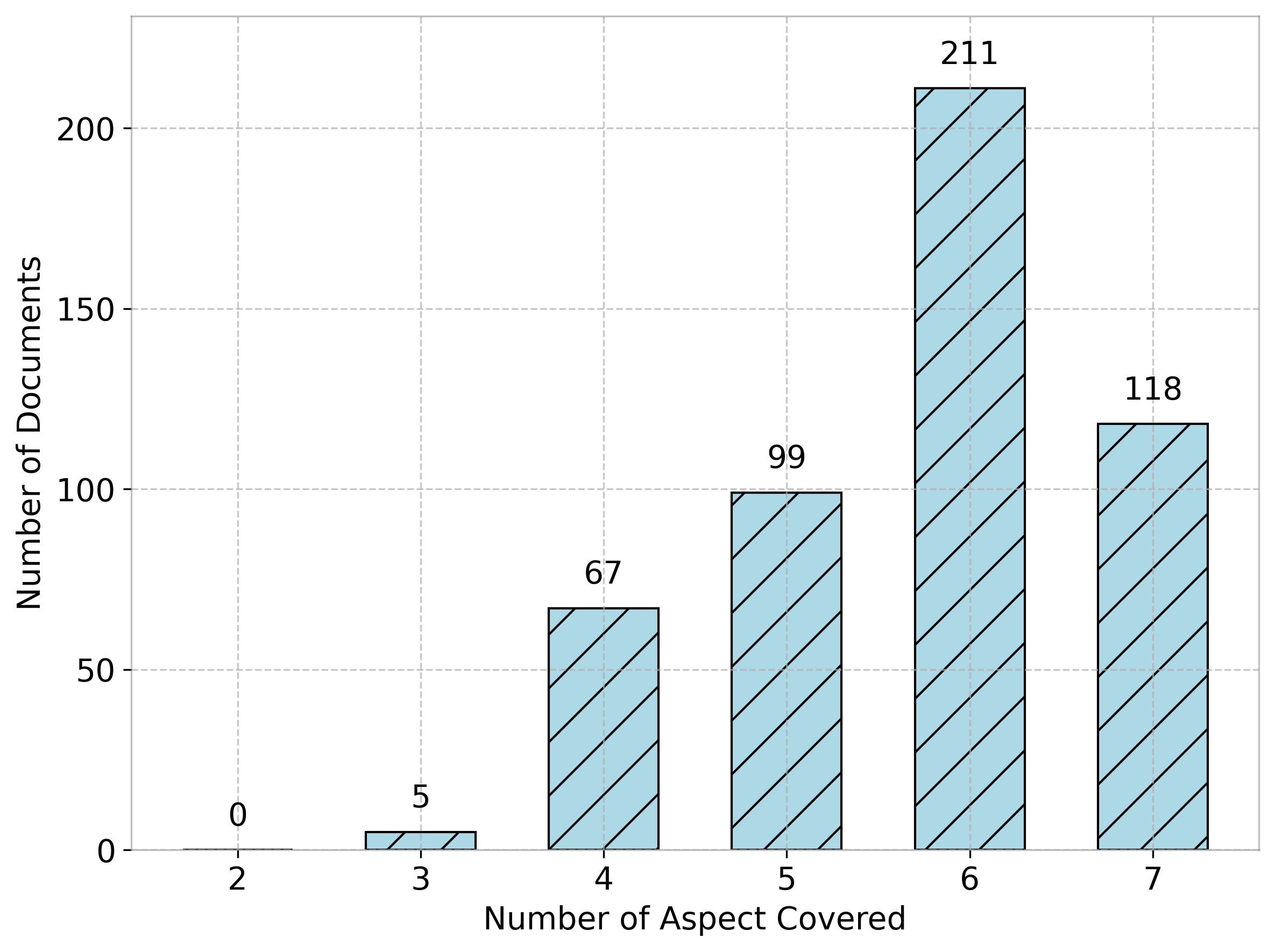}
            \caption{Aspect coverage across abstracts.}
            \label{fig:aspect_distribution}
        \end{subfigure}
        \begin{subfigure}[b]{0.245\textwidth}
            \includegraphics[width=\linewidth]{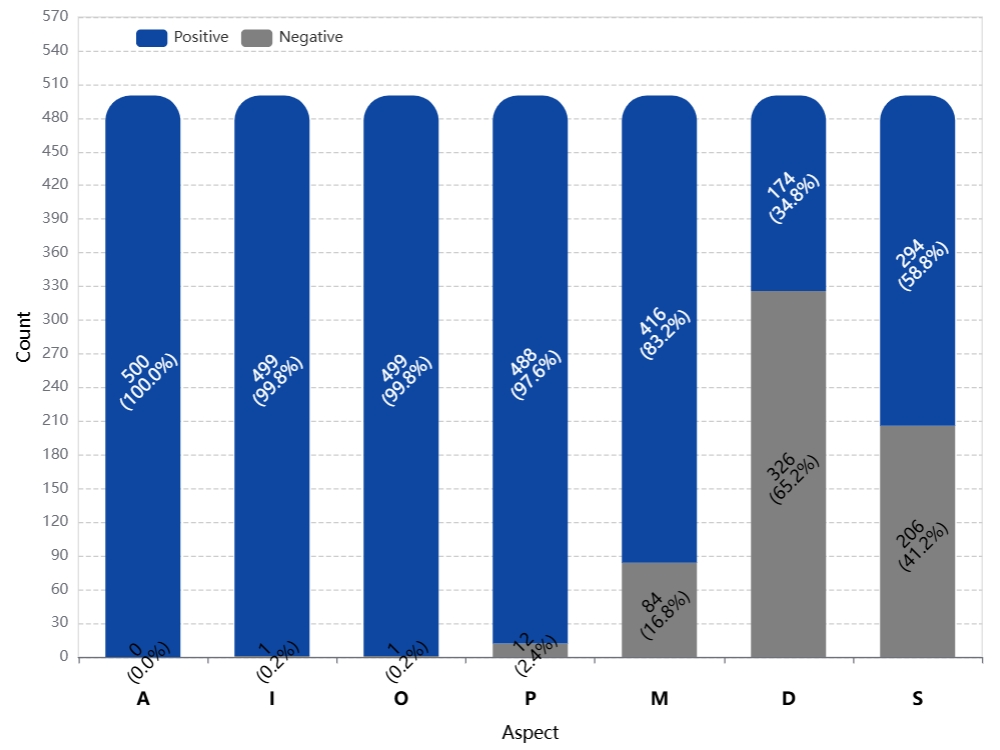}
            \caption{Proportion of positive and negative data.}
            \label{fig:positive_negative}
        \end{subfigure}
        \caption{Characteristics of the \textsc{TracSum}dataset.}
        \vspace{-15pt}
    \end{figure*}

    \subsection{Source Article Length} 
    Among the 500 abstracts, the average length per abstract was 319.89 tokens, with the longest containing 1,104 tokens and the shortest containing only 25. The distribution of token counts across abstracts is illustrated in \autoref{fig:tokens_distribution}.  Additionally, each abstract contained an average of 10.42 sentences, with sentence counts ranging from 1 to 32. The distribution of sentence counts is shown in \autoref{fig:sentence_distribution}.

    \subsection{Aspect Coverage in Abstracts}
    All 500 documents contained information on at least three aspects. Among them, 118 documents covered all seven aspects, and 211 documents covered six aspects, as illustrated in \autoref{fig:aspect_distribution}.

    \subsection{Proportion of Positive and Negative Data}
    We analyzed the distribution of positive and negative data samples across seven aspects, as shown in \autoref{fig:positive_negative}. All 500 abstracts included aspect A (Research Aims), while 499 covered aspect I (Research Methods or Intervention) and aspect O (Research Results or Outcomes). In contrast, aspect D (Treatment Duration) was less common, appearing in only 174 abstracts. Overall, the ratio of positive to negative samples was 2862:638.

    \subsection{Length of Traceable Summaries}
    As shown in \autoref{tab:size_sum}, all 2,862 positive summaries had an average length of 28.06 tokens, with the longest containing 77 tokens and the shortest just 3. On average, each summary cited 1.78 sentences, with the number ranging from 1 to 7. Among all aspects, summaries related to aspect S (Side Effects) had the highest average token count, while those concerning aspect I (Research Methods or Intervention) cited the most sentences.

    \begin{table*}[t]
        \small
        \centering
        \begin{tabularx}{\textwidth}{c c c c c c c c | c c c c c c c}
        \toprule
        \multicolumn{8}{c|}{\textit{\textbf{Summary}}} & \multicolumn{7}{c}{\textit{\textbf{Citations}}} \\ 
        \cmidrule(lr){2-8} \cmidrule(lr){9-14}
        \textbf{} & \textbf{A} & \textbf{I} & \textbf{O} & \textbf{P} & \textbf{M} & \textbf{D} & \textbf{S} & \textbf{A} & \textbf{I} & \textbf{O} & \textbf{P} & \textbf{M} & \textbf{D} & \textbf{S} \\ 
        \midrule
        Min & 13 & 15 & 12 & 4 & 3 & 4 & 4 & 1 & 1 & 1 & 1 & 1 & 1 & 1 \\

        Max & 56 & 73 & 77 & 69 & 77 & 75 & 75 & 5 & 7 & 6 & 5 & 6 & 5 & 4 \\

        Avg. & 29.33 & 37.81 & 34.75 & 25.64 & 25.37 & 17.82 & 25.67 & 1.51 & 2.33 & 2.58 & 1.61 & 1.74 & 1.25 & 1.46 \\
        \bottomrule
        \end{tabularx}
        \caption{Length of summaries (in tokens) and number of citations (in sentences) in positive samples.}
        \label{tab:size_sum}
    \end{table*}

\section{Generation Pipelines} \label{sec:generation_pipelines}
    In this section, we provide a detailed description of the design and training of our three baseline methods: \textsc{Track-Then-Sum}, \textsc{Sum-Then-Track}, and \textsc{End-To-End}.
    
    \subsection{\textsc{Track-Then-Sum}} \label{sec:tts}
\begin{figure*}[t]
        \includegraphics[width=\linewidth]{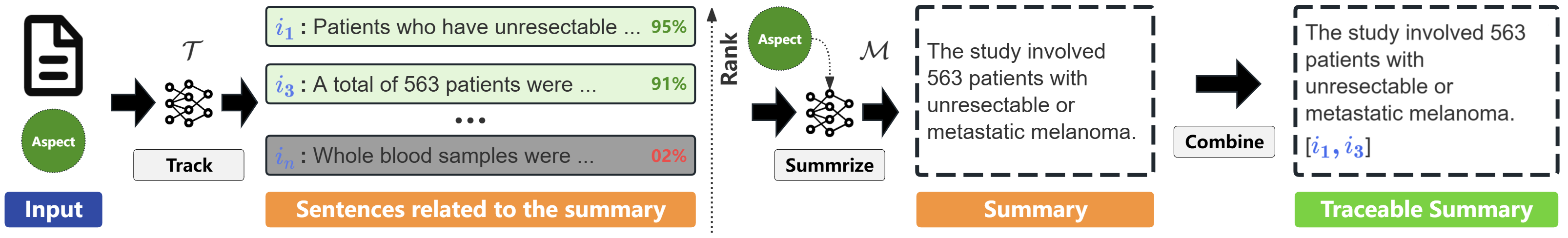}
        \caption{\textsc{Track-Then-Sum} summarization pipeline.}
        \label{fig:Track-Then-Sum}
        \vspace{-15pt}
    \end{figure*}

    As illustrated in \autoref{fig:Track-Then-Sum}, the \textsc{Track-Then-Sum} generation pipeline consists of two phases: tracking and summarization. In the first phase, the tracker module $\mathcal{T}$ retrieves the sentences most relevant to the given aspect using a default threshold of 0.5. In the second phase, the summarizer module $\mathcal{S}$ generates a concise summary based on the selected sentences. Finally, the summary and the cited sentences are merged to form the final system output.

    \subsection{Tracker $\mathcal{T}$}
    We implement the sentence tracing task as a binary classification of sentences within the abstract.

    \noindent\textbf{Data Collection:} We applied sentence tokenization to each abstract in the training set. For every sentence, we created \((c, a)\) pairs by combining it with each predefined aspect $a \in \mathcal{A}$. Each pair was labeled with a binary variable \(y\) based on the corresponding \textit{citations} field: if the sentence index appeared in the \textit{citations} associated with aspect $a$, we assigned \(y=1\); otherwise, \(y=0\). In total, we obtained 35.5K sentence-aspect-label pairs, forming the training dataset $\mathcal{D}_{\mathcal{T}}$. 
    
    \vspace{0.15cm}
    
    \noindent\textbf{Training:} Given the constructed dataset $\mathcal{D}_{\mathcal{T}}$, we initialized tracker \(\mathcal{T}\) using a pre-trained language model (LM) as the backbone. The model was subsequently fine-tuned on \(\mathcal{D}_{\mathcal{T}}\) using a standard binary classification objective which maximizes the log-likelihood of the observed labels: 
    \begin{equation*}
    \max_{\mathcal{T}} \mathbb{E}_{((c, a),y)\sim\mathcal{D}_{\mathcal{T}}} \log p_{\mathcal{T}}(y \mid (c, a))
    \end{equation*}
    We fine-tuned the tracker $\mathcal{T}$ using the QLoRA technique, initializing from the 4-bit quantized version of the LLaMA-3.1-8B-Instruct backbone\footnote{Model: meta-llama/Llama-3.1-8B}, on $\mathcal{D}_{\mathcal{T}}$. To enable binary classification, we appended a lightweight classification head that maps the model's output to a single scalar representing the predicted probability. Training was conducted on six NVIDIA A6000 GPUs with a batch size of 32, gradient accumulation steps of 2, and a total of 5 epochs. We employed a learning rate of $1 \times 10^{-5}$, applied a weight decay of 0.01, set the random seed to 3407 for reproducibility, and used 200 warmup steps. The full training process took 17 hours and 2 minutes.

    \subsection{Summarizer $\mathcal{S}$}
    
    \textbf{Data Collection:}  For each summary $sum$ in the training set, we extracted related sentences from the abstract based on the \textit{citations} field to form the set $\mathcal{C}$. Each $\mathcal{C}$ was paired with its associated aspect $a$, and combined with the $sum$ to form $((\mathcal{C}, a), sum)$. In total, we obtained 2.8K citations-aspect-summary pairs, forming the training dataset $\mathcal{D}_{\mathcal{S}}$. 
    
    \vspace{0.15cm}
    
    \noindent\textbf{Training:} Similar to the training of $\mathcal{T}$, we initialized summarizer $\mathcal{S}$ using a pre-trained LM as the backbone. We then fine-tuned summarizer $\mathcal{S}$ on $\mathcal{D}_{\mathcal{S}}$ using a standard next-token prediction objective, which maximizes the likelihood of generating the target summary $sum$ given the input $(\mathcal{C}, a)$ pair:
    \begin{equation*}
        \max_{\mathcal{S}} \mathbb{E}_{((C, a),sum)\sim\mathcal{D}_{\mathcal{S}}} \log p_{\mathcal{S}}(sum \mid C, a)
    \end{equation*}
    The input instruction is shown in \autoref{tab:prompt_tts}. We fine-tuned Summarizer $\mathcal{S}$ using the Unsloth framework, starting from the 4-bit version of the LLaMA-3.1-8B-Instruct base model\footnote{Model: unsloth/Meta-Llama-3.1-8B-Instruct-bnb-4bit}, on $\mathcal{D}_{\mathcal{S}}$. Training was performed on two NVIDIA A6000 GPUs with a batch size of 16, a gradient accumulation step size of 2, and a total of 5 epochs. We used a learning rate of 1e-5, a weight decay of 0.01, a fixed random seed of 3407, and 200 warmup steps. The entire training process took 1 hour and 55 minutes. Additionally, we adopted the \texttt{train\_on\_responses\_only} strategy to focus learning on relevant output segments.
    \begin{figure*}[t]
        \includegraphics[width=\linewidth]{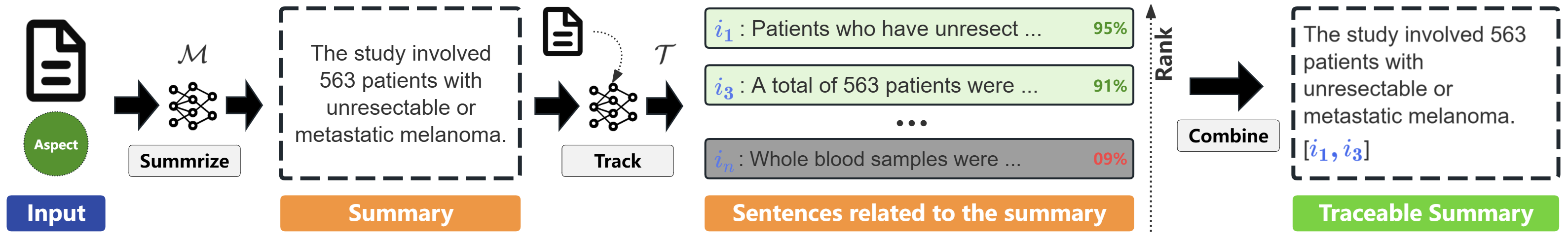}
        \caption{\textsc{Sum-Then-Track} method pipeline.}
        \label{fig:Sum-Then-Track}
        \vspace{-15pt}
    \end{figure*}

    \subsection{TTS $\oplus \ f.$}
    As mentioned in \S\ref{sec:baseline}, our \textsc{Track-Then-Sum} method includes two variants, differing only in their input. Specifically, the TTS $\oplus \ f.$ variant uses both the set of cited sentences and the full context (i.e., abstract) as input. The input instruction is shown in \autoref{tab:prompt_tts_f}. All other settings remain unchanged, except for the batch size, which was set to 8. Under this configuration, training took 8 hours and 36 minutes.
    
    \subsection{\textsc{Sum-Then-Track}}
    \subsubsection{Inference Overview}
    As illustrated in \autoref{fig:Sum-Then-Track}, the \textsc{Sum-Then-Track} method consists of two phases: summarization and tracking. In the first phase, the summarizer $\mathcal{S}$ generates an aspect-specific summary $sum$ from an abstract $d$ based on a given aspect $a$. In the second phase, the tracker $\mathcal{T}$ identifies the sentences most relevant to this summary using a default similarity threshold of 0.5. Finally, the summary and the corresponding sentences are combined to form the final output, as shown in \hyperref[tab:algorithm3]{Algorithm 3}.

    \begin{table}[t]
    \small
    \begin{tabularx}{0.49\textwidth}{l}
    \toprule
    \textbf{Algorithm 3:} \textsc{Sum-Then-Track} Inference \\ 
    \midrule
    \makecell[l]{\textbf{Require:} Tracker $\mathcal{T}$, Summarizer $\mathcal{S}$} \\ 
    
    \textbf{Input:} article $d=\{c_{1}, c_{2}, ...,c_{n}\}$ and aspect $a \in \mathcal{A}$  \\ 
    \textbf{Output:} summary $sum$ and its citations $\mathcal{C}'$ \\

    1: $sum \leftarrow \mathcal{S}(a, d)$; \\
    2: $\mathcal{C}' \leftarrow \emptyset$; \\
    3: \textbf{\textcolor{myBlue}{foreach}} $c_i \in \{c_{1}, c_{2}, ...,c_{n}\}$  \\
    4: \ \ \ \ \ \ $\mathcal{T}$ predict \textbf{\textcolor{myOrange}{relevance}} given $(sum, c_i)$; \\
    5: \ \ \  \ \ \ \textbf{\textcolor{myBlue}{if}} \textbf{\textcolor{myOrange}{relevance}} == Yes \textbf{\textcolor{myBlue}{then}} append $c$ to $\mathcal{C}'$; \\
    
    \bottomrule
    \end{tabularx}
    \caption*{Algorithm 3: \textsc{Sum-Then-Track} inference process. }
    \vspace{-15pt}
    \end{table}\label{tab:algorithm3}

    \subsubsection{Summarizer $\mathcal{S}$}
    \textbf{Data Collection:} We extracted abstract, aspect, and summary fields from the training set, resulting in 2.8K ((\(d\), \(a\)), \(sum\)) pairs, denoted as $\mathcal{D}_{\mathcal{S}}$.

    \vspace{0.15cm}
    
    \noindent\textbf{Training:} We then initialized summarizer $\mathcal{S}$ using a pre-trained LM as the backbone. We then fine-tuned summarizer $\mathcal{S}$ on $\mathcal{D}_{\mathcal{S}}$ using a standard next-token prediction objective, which maximizes the likelihood of generating the target summary $sum$ given the input $(d, a)$ pair:
    \begin{equation*}
        \max_{\mathcal{S}} \mathbb{E}_{((d, a),sum)\sim\mathcal{D}_{\mathcal{S}}} \log p_{\mathcal{S}}(sum \mid d, a)
    \end{equation*}
     The input instruction is shown in \autoref{tab:prompt_stt}. We fine-tuned Summarizer $\mathcal{S}$ using the Unsloth framework, starting from the 4-bit version of the LLaMA-3.1-8B-Instruct base model, on $\mathcal{D}_{\mathcal{S}}$. Training was performed on two NVIDIA A6000 GPUs with a batch size of 8, a gradient accumulation step size of 2, and a total of 5 epochs. We used a learning rate of 1e-5, a weight decay of 0.01, a fixed random seed of 3407, and 200 warmup steps. The entire training process took 7 hour and 32 minutes. Additionally, we adopted the \texttt{train\_on\_responses\_only} strategy to focus learning on relevant output segments.

    \subsubsection{Tracker $\mathcal{T}$}
    \textbf{Data Collection:} We first applied sentence tokenization to all abstracts in the training set. For each abstract, every sentence $c$ was paired with each summary $sum$, forming ($c$, $sum$) pairs. Each pair was then labeled with $y$ based on the \textit{citations} field. This process resulted in 35.5k (($c$, $sum$), $y$) pairs, denoted as $\mathcal{D}_{\mathcal{T}}$. 

    \vspace{0.15cm}

    \noindent\textbf{Training:} Given the constructed dataset $\mathcal{D}_{\mathcal{T}}$, we initialized tracker \(\mathcal{T}\) using a pre-trained language model (LM) as the backbone. The model was subsequently fine-tuned on \(\mathcal{D}_{\mathcal{T}}\) using a standard binary classification objective which maximizes the log-likelihood of the observed labels: 
    \begin{equation*}
    \max_{\mathcal{T}} \mathbb{E}_{((c, sum),y)\sim\mathcal{D}_{\mathcal{T}}} \log p_{\mathcal{T}}(y \mid (c, sum))
    \end{equation*}
    We fine-tuned the tracker $\mathcal{T}$ using the QLoRA technique, initializing from the 4-bit quantized version of the LLaMA-3.1-8B-Instruct backbone, on $\mathcal{D}_{\mathcal{T}}$. To enable binary classification, we appended a lightweight classification head that maps the model's output to a single scalar representing the predicted probability. Training was conducted on six NVIDIA A6000 GPUs with a batch size of 32, gradient accumulation steps of 2, and a total of 5 epochs. We employed a learning rate of $1 \times 10^{-5}$, applied a weight decay of 0.01, set the random seed to 3407 for reproducibility, and used 200 warmup steps. The full training process took 22 hours and 12 minutes.

    \begin{table*}[h]
        \small
        \begin{tabularx}{\linewidth}{l c c c c c c}
        \toprule
        \textbf{Model} & \textbf{API Src.} & \textbf{Input Prices} & \textbf{Output Prices} & \textbf{Input Length} & \textbf{Output Length} & \textbf{Costs} \\
        \midrule
    
        Llama-3.1-8B-Inst. & DeepInfra & \$0.03 & \$0.05 & 131K & 8K & \$0.030 \\
    
        Llama-3.3-70B-Inst. & DeepInfra & \$0.23 & \$0.40 & 131K & 8K & \$0.250 \\
    
        Mistral-7B-Inst (V0.3). & DeepInfra & \$0.029 & \$0.055 & 32K & 8K & \$0.040 \\
    
        Mistral-8x7B-Inst. & DeepInfra & \$0.24 & \$0.24 & 131K & 4K & \$0.600 \\
        
        Gemma-3-12B-Inst. & DeepInfra & \$0.05 & \$0.100 & 128K & 8K & \$0.070 \\
        
        Gemma-3-27B-Inst. & DeepInfra & \$0.10 & \$0.20 & 128K & 8K & \$0.110 \\
    
        GPT-4o & OpenAI & \$2.50 & \$10.0 & 128K & 16K & \$2.838 \\
        
        GPT-4o-mini & OpenAI & \$0.15 & \$0.60 & 128K & 16K & \$0.147 \\
    
        \midrule
        & & & & & \multicolumn{2}{r}{SUM : \$4.085}\\
        \bottomrule
        \end{tabularx}
        \caption{Details on the use of different model APIs.}
        \label{tab:api_cost}
        \vspace{-15pt}
    \end{table*}
    \subsection{\textsc{End-To-End}}
\begin{figure}[t] 
        \includegraphics[width=\linewidth]{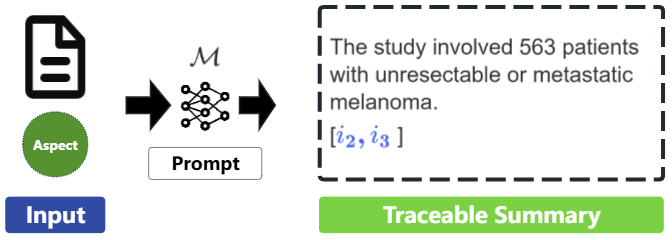}
        \caption{\textsc{End-to-End} generation pipeline.}
        \label{fig:ent-to-end}
    \end{figure}

    The \textsc{End-to-End} approach employs a single model $\mathcal{M}$, to jointly perform summarization and sentence tracking, as shown in \autoref{fig:ent-to-end}.
    
    \subsubsection{Inference Phase}
    
    Given a abstract $d$ and an aspect $a \in \mathcal{A}$, $\mathcal{M}$ generates a summary focused on $a$ and $\mathcal{C}'$ on which the summary relies, as illustrated in \hyperref[tab:algorithm4]{Algorithm 4}.

    \begin{table}[t]
    \small
    \begin{tabularx}{0.49\textwidth}{l}
    \toprule
    \textbf{Algorithm 4:} \textsc{End-To-End} Inference \\ 
    \midrule
    \makecell[l]{\textbf{Require:} Model $\mathcal{M}$} \\ 
    
    \textbf{Input:} article $d=\{c_{1}, c_{2}, ...,c_{n}\}$ and aspect $a \in \mathcal{A}$  \\ 
    \textbf{Output:} summary $sum$ and its citations $\mathcal{C}'$ \\
    1: $\mathcal{C}' \leftarrow \emptyset$; \\
    2: $(sum, \mathcal{C}') \leftarrow \mathcal{M}(a, d)$; \\
    \bottomrule
    \end{tabularx}
    \caption*{Algorithm 4: \textsc{End-To-End} inference process. }
    \vspace{-15pt}
    \end{table}\label{tab:algorithm4}
    
    \subsubsection{Training Phase}
    
    \textbf{Data Collection.} We extracted \textit{abstract}, \textit{aspect}, \textit{summary}, and \textit{citations} fields from the training set and then combined them into (($d$, $a$), ($sum$, $\mathcal{C}$)) pairs. As a result, we obtained 2.8K training instances, denoted by $\mathcal{D}_{\mathcal{M}}$.

    \vspace{0.15cm}
    
    \noindent\textbf{Training.} We then initialized $\mathcal{M}$ with a pre-trained LM and trained it on $\mathcal{D}_{\mathcal{M}}$ using a standard conditional language modeling objective, maximizing the likelihood: 
    \begin{equation*}
        \max_{\mathcal{M}} \mathbb{E}_{((d, a), (sum, \mathcal{C})) \sim \mathcal{D}_{\mathcal{M}}} \log p_{\mathcal{M}}(sum, \mathcal{C} \mid d, a)
    \end{equation*}
    The input instruction is shown in \autoref{tab:prompt_ete}. We fine-tuned $\mathcal{M}$ using the Unsloth framework, starting from the 4-bit version of the LLaMA-3.1-8B-Instruct base model, on $\mathcal{D}_{\mathcal{M}}$. Training was performed on two NVIDIA A6000 GPUs with a batch size of 8, a gradient accumulation step size of 2, and a total of 5 epochs. We used a learning rate of 1e-5, a weight decay of 0.01, a fixed random seed of 3407, and 200 warmup steps. The entire training process took 8 hours and 16 minutes. Additionally, we adopted the \texttt{train\_on\_responses\_only} strategy to focus learning on relevant output segments.


\section{API Cost} \label{sec:api_cost}
\subsection{Dataset Collection Costs}
We initially generated our dataset with the free credits provided by the Mistral-Large API, so the cost for this part is \$0. 

\subsection{Evaluation Costs}
We incurred approximately \$4.085 in API costs to obtain results from eight different models on the test set, as detailed in \autoref{tab:api_cost}. The test set comprises 700 data samples, each formatted into prompts, resulting in approximately 100K input tokens in total. The number of output tokens varies across LLMs; standard text generation models typically produce around 50K output tokens.

\section{Experiment Analysis} 
\subsection{Full Context $\oplus \ \mathcal{C}$ vs. $\mathcal{C}$ only} \label{sec:full_analysis}
In this section, we present an example to illustrate how incorporating full context impacts summary generation and, in turn, affects claim recall. When the cited sentences (i.e., the tracker $\mathcal{T}$ output) remain fixed, providing the full document as additional input enables the summarizer $\mathcal{S}$ to better resolve abbreviations and domain-specific terminology, thereby enhancing claim recall. As shown in \autoref{tab:full_analysis}, TTS $\oplus f$ resolves the abbreviation ``RT'' as ``radiation therapy'', which leads the NLI model (TRUE) to determine that the subclaim is entailed by the reference text during entailment evaluation. This results in an increase in the overall claim recall score from 2/4 to 3/4.

However, providing additional context beyond the cited sentences may cause the summarizer $\mathcal{S}$ to incorporate irrelevant or unsupported information (i.e., content not present in the cited sentences), which could reduce claim precision or citation-based metrics. Nonetheless, our evaluation results do not show a noticeable drop in other metrics. This may be attributed to the instruction explicitly directing the summarizer $\mathcal{S}$ to generate summaries strictly based on the cited sentences, with the additional context serving only as reference.

\begin{table}[t]
    \small
    \centering
    \begin{tabularx}{0.48\textwidth}{X}
        \toprule
        \rowcolor{gray!20} 
        \textbf{Reference:} \quad Summary \tikz[baseline=-0.5ex] \draw[->, thick] (0,0) -- (0.3,0) -- (0.3,-0.2); \quad Citations $\rightarrow$ 0, 1, 7 \\
        
        1$'$. A total of 50 participants were involved in the study. \\ 
        2$'$. Participants with cutaneous neurotropic melanoma of the head and neck. \\
        3$'$. 23 participants were assigned to the observation group. \\
        4$'$. 27 participants were assigned to the radiation therapy group. \\[0.1cm]
        \hdashline
        \scriptsize \textcolor{myBlue}{Citation 0:} BACKGROUND: Cutaneous neurotropic melanoma (NM) of the head and neck (H\&N) is prone to local relapse, possibly due to difficulties widely excising the tumor.
        \scriptsize \textcolor{myBlue}{Citation 1:} This trial assessed radiation therapy (RT) to the primary site after local excision.
        \scriptsize \textcolor{myBlue}{Citation 7:} During 2009-2020, 50 participants were randomized: 23 to observation and 27 to RT. \\
        
        \midrule
        \rowcolor{gray!20} 
        \textbf{TTS Output:} \quad Subclaims \tikz[baseline=-0.5ex] \draw[->, thick] (0,0) -- (0.3,0) -- (0.3,-0.2); \quad Citations $\rightarrow$ 7 \\
        1$'$. A total of 50 participants were randomized in the study. \\ 
        2$'$. 23 participants were assigned to the observation group. \\
        3$'$. 27 participants were assigned to the RT group. \\[0.1cm]
        \hdashline
        (TRUE) Claim Recall: 2/4. 1$'$: \textcolor{green}{\textbf{\ding{51}}}, 2$'$: \textcolor{green}{\textbf{\ding{51}}}, 3$'$: \textcolor{red}{\textbf{\ding{55}}} \\
        \midrule
        \rowcolor{gray!20} 
        \textbf{TTS $\oplus \ f.$ Output:} \quad Subclaims \tikz[baseline=-0.5ex] \draw[->, thick] (0,0) -- (0.3,0) -- (0.3,-0.2); \quad Citations $\rightarrow$ 7 \\
        
        1$'$. A total of 50 participants were randomized in the study. \\ 
        2$'$. 23 participants were assigned to the observation group. \\
        3$'$. 27 participants were assigned to the radiation therapy (RT) group. \\[0.1cm]
        \hdashline
        (TRUE) Claim Recall: 3/4. 1$'$: \textcolor{green}{\textbf{\ding{51}}}, 2$'$: \textcolor{green}{\textbf{\ding{51}}}, 3$'$: \textcolor{green}{\textbf{\ding{51}}} \\
        \bottomrule
    \end{tabularx}
    \caption{An example of summaries generated by TTS and TTS $\oplus f$, along with their claim recall comparison (PMID: 38851639, Aspect: Patients).}
    \vspace{-15pt}
    \label{tab:full_analysis}
\end{table}

\subsection{Agreement with Human Evaluation} \label{sec:agreement_with_human}
To evaluate the relationship between the system outputs and task-level evaluation scores, we employ both \textit{Spearman’s correlation coefficient} ($\rho$) \cite{kendall1990rank} and \textit{Pearson’s correlation coefficient} ($r$) \cite{sheskin2003handbook}. Pearson’s $r$ measures the strength of a \textit{linear relationship} between two continuous variables, which is appropriate when assuming interval-scaled outputs and normally distributed scores \cite{benesty2009pearson}. In contrast, Spearman’s $\rho$ captures \textit{monotonic relationships} based on rank order, making it more robust to non-linear patterns and outliers \cite{hauke2011comparison}. Using both metrics provides a comprehensive view of how well the automatic system outputs align with human-centric evaluation criteria, accounting for both linear trends and ordinal consistency.

\begin{figure}[t]
    \centering
    \includegraphics[width=\linewidth]{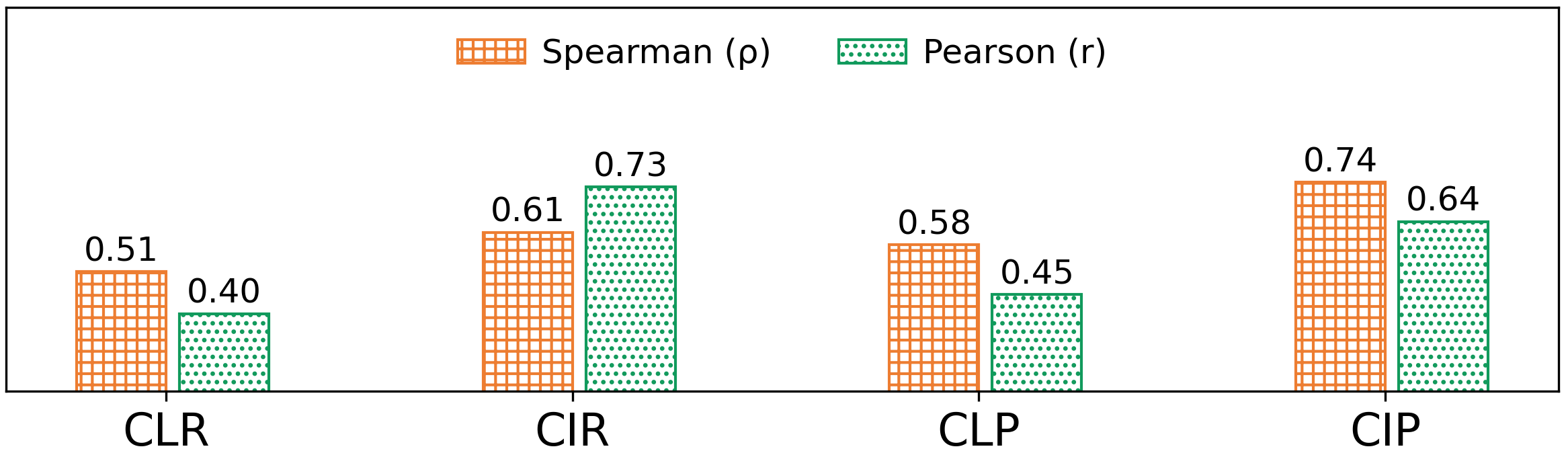}
   \caption{Spearman’s correlation coefficient ($\rho$) and Pearson’s correlation coefficient ($r$) between TRUE and human evaluation scores across four evaluation metrics.}
  \label{fig:human_agreement}
  \vspace{-15pt}
\end{figure}

Specifically, we randomly sampled ten abstracts from the test set, and asked the annotator to follow the procedure in \hyperref[tab:algorithm2]{Algorithm 2} to assess outputs from the best-performing method (TTS~$\oplus \ f.$) using four evaluation metrics. As indicated in \autoref{tab:agreement_with_human}, human evaluations score higher than the TRUE model on most metrics, achieving an F1 score of 74.3 for claims and 76.2 for citations quality. For each of the four evaluation metrics, we computed the Spearman correlation coefficient ($\rho$) and Pearson correlation coefficient ($r$)  between the automatic evaluation results and human judgments. As shown in \autoref{fig:human_agreement}, the Spearman correlation coefficient between human and automatic evaluation results is $\rho = 0.612$, and the Pearson correlation coefficient is $r = 0.577$. The agreement is relatively lower for claim-related metrics, whereas citation-related metrics demonstrate stronger consistency with human judgments.

\begin{table}[t]
    \small
    \centering
    \begin{tabularx}{0.48\textwidth}{
        >{\hspace{-3pt}\centering\arraybackslash\hspace{-3pt}}c :
        >{\hspace{-3pt}\centering\arraybackslash}c
        >{\hspace{-3pt}\centering\arraybackslash}c :
        >{\hspace{-3pt}\centering\arraybackslash}c
        >{\hspace{-3pt}\centering\arraybackslash}c :
        >{\hspace{-3pt}\centering\arraybackslash}c
        >{\hspace{-3pt}\centering\arraybackslash}c
        }
        \toprule
   
        & \multicolumn{2}{c}{\textbf{\textcolor{myBlue}{{\scriptsize Completeness}}}} & \multicolumn{2}{c}{\textbf{\textcolor{myOrange}{{\scriptsize Conciseness}}}} & \multicolumn{2}{c}{\textbf{{\scriptsize F1 Score}}} \\
        \cmidrule(lr){2-3} \cmidrule(lr){4-5} \cmidrule(lr){6-7}
        \textbf{Evaluator} & \textbf{CLR} & \textbf{CIR} & \textbf{CLP} & \textbf{CIP} & $F_1^{\text{cl.}}$ & $F_1^{\text{ci.}}$ \\
        \midrule
        Human & 81.1$\uparrow$ & 74.3$\uparrow$ & 68.6$\uparrow$ & 78.1$\downarrow$ & 74.3$\uparrow$ & 76.2$\downarrow$ \\
        TRUE & 78.2 & 73.4 & 65.7 & 79.5 & 71.4 &  76.3 \\

        \bottomrule
     \end{tabularx}
    \caption{Comparison of evaluation results between human annotator and the TRUE model on 10 sampled abstracts.}
    \label{tab:agreement_with_human}
    \vspace{-10pt}

\end{table}

\subsection{Comparison of Entailment Evaluators} \label{sec:comparison_evaluator}
We experiment with two additional instruction-following LLMs as entailment evaluators: the proprietary GPT-4o \cite{hurst2024gpt} and the open-source Mistral-Large \cite{mistral2025}. Building on the experimental setup described in \S\ref{sec:huamn_agreement}, we replace the TRUE model with each of these evaluators to assess the outputs generated by the TTS~$\oplus~f.$ variant. The evaluation results are presented in \autoref{tab:evaluators}. Among the models, GPT-4o produces scores that most closely align with human judgments, followed by Mistral.
\begin{table}[h]
    \small
    \centering
    \begin{tabularx}{0.48\textwidth}{
        >{\centering\arraybackslash}c :
        >{\centering\arraybackslash}c
        >{\centering\arraybackslash}c :
        >{\centering\arraybackslash}c
        >{\centering\arraybackslash}c :
        >{\centering\arraybackslash}c
        >{\centering\arraybackslash}c
        }
        \toprule
   
        & \multicolumn{2}{c}{\textbf{\textcolor{myBlue}{{Completeness}}}} & \multicolumn{2}{c}{\textbf{\textcolor{myOrange}{{Conciseness}}}} & \multicolumn{2}{c}{\textbf{{F1 Score}}} \\
        \cmidrule(lr){2-3} \cmidrule(lr){4-5} \cmidrule(lr){6-7}
        \textbf{Evaluator} & \textbf{CLR} & \textbf{CIR} & \textbf{CLP} & \textbf{CIP} & $F_1^{\text{cl.}}$ & $F_1^{\text{ci.}}$ \\
        \midrule
        Human & 81.1 & 74.3 & 68.6 & 78.1 & 74.3 & 76.2 \\
        TRUE & 78.2 & 73.4 & 65.7 & 79.5 & 71.4 &  76.3 \\
        GPT-4o & 80.2 & 77.1 & 67.0 & 76.2 & 73.0 &  76.7 \\
        Mistral & 75.6 & 76.8 & 70.1 & 74.5 & 72.8 &  75.6 \\

        \bottomrule
     \end{tabularx}
    \caption{Comparison of evaluation results between human annotator and three entailment evaluators on 10 sampled abstracts.}
    \label{tab:evaluators}
    \vspace{-10pt}

\end{table}

\clearpage
\onecolumn
\section{Data Samples of \textsc{TracSum} Dataset} \label{sec:data_sample}
\vspace{0.5em}
\small
\begin{center}
\begin{tabularx}{\textwidth}{m{0.08\textwidth}  m{0.3\textwidth}  m{0.06\textwidth} m{0.28\textwidth}  m{0.08\textwidth}  m{0.03\textwidth}}

    \toprule
    \textbf{PMID} & \textbf{abstract} & \textbf{aspect} & \textbf{summary} & \textbf{citations} & ... \\
    
    \midrule
    31638282 & The multinational phase 3 CheckMate 238 trial compared adjuvant therapy with nivolumab versus ipilimumab among patients with resected stage III or IV melanoma (N = 906)... & \centering d & Unknown. & \centering [] & ... \\
    
    \midrule
    33294860 & In this study, we incorporate analyses of genome-wide sequence and structural alterations with pre- and on-therapy transcriptomic and T cell repertoire features in immunotherapy-naive melanoma patients treated with ... & \centering a & The study aims to predict response to immune checkpoint blockade by integrating genomic, transcriptomic, and immune repertoire data. & \centering [ 4 ] & ... \\

    \midrule
    34650833 & Combination immunotherapy with sequential administration may enhance metastatic melanoma (MM) patients with long-term disease control. High Dose Aldesleukin/Recombinant Interleukin-2 (HD rIL-2) and ipilimumab (IPI) offer... & \centering m & The study used High Dose Aldesleukin/Recombinant Interleukin-2 (HD rIL-2) at 600,000 IU/kg and ipilimumab (IPI) at 3 mg/kg. & \centering [ 1, 3 ] & ... \\

    \midrule
    37479483 & BACKGROUND: Continuous combination of MAPK pathway inhibition (MAPKi) and anti-programmed death-(ligand) 1 (PD-(L)1) showed high response rates, but only limited improvement in progression-free survival (PFS) at the cost of a high frequency... & \centering p & The study involved 33 patients with treatment-naïve BRAFV600E/K-mutant advanced melanoma, with 32 randomized into four cohorts. & \centering [ 3, 8 ] & ... \\

    \midrule
    33593880 & PURPOSE: Triple-negative breast cancer (TNBC) is an aggressive disease with limited therapeutic options. Antibodies targeting programmed cell death protein 1 (PD-1)/PD-1 ligand 1 (PD-L1) have entered the therapeutic landscape in TNBC, but only a minority of patients benefit. A way to reliably enhance immunogenicity, T-cell infiltration, and predict responsiveness is critically needed. PATIENTS AND METHODS: Using mouse models of TNBC... & \centering i & This study used mouse models of TNBC to evaluate immune activation and tumor targeting of intratumoral IL12 plasmid followed by electroporation (Tavo), conducted a single-arm prospective clinical trial of Tavo monotherapy in patients with treatment-refractory advanced TNBC, and expanded findings using publicly available breast cancer and melanoma datasets. & \centering [ 3, 4, 5 ] & ... \\

    \midrule
    38870745 & BACKGROUND: Treatment options for immunotherapy-refractory melanoma are an unmet need. The MASTERKEY-115 phase II, open-label, multicenter trial evaluated talimogene ... & \centering s & Treatment-related adverse events (TRAEs), including grade $\geq$3 TRAEs, serious AEs, and fatal AEs, occurred in 76.1\%, 12.7\%, 33.8\%, and 14.1\% of patients, respectively. & \centering [ 11 ] & ... \\

    \midrule
    33127652 & PURPOSE: Increased $\beta$-adrenergic receptor ($\beta$-AR) signaling has been shown to promote the creation of an immunosuppressive tumor microenvironment (TME) ... & \centering o & The combination of propranolol with pembrolizumab in treatment-naïve metastatic melanoma is safe and shows very promising activity with an objective response rate of 78\%. & \centering [ 12,14 ] & ... \\
    \bottomrule
    \end{tabularx}
    \captionof{table}{Seven traceable aspect-based summary samples from \textsc{TracSum} dataset.} \label{tab:data_sample}
\end{center}

\clearpage
\section{Instructions} \label{sec:llm_prompt}

\subsection{LLM Prompt Template}
\begin{center}
    \small
    \begin{tabular}{p{\textwidth}}
    \toprule
    \textbf{Instructions} \\
    Given a document consisting of a set of sentences with a marker attached to the head of each sentence. Based on the demonstrations, please summarize the \uwave{research questions or aims} of this study in one sentence and output the sentence markers involved. If there is no relevant information in the document, answer \textcolor{gray}{"Unknown"}.\\
    \textbf{Document} \\
    `[
        "\textcolor{blue}{0}: The EORTC-STBSG coordinated two large trials of adjuvant chemotherapy (CT) in localized high-grade soft tissue sarcoma (STS).",
        "\textcolor{blue}{1}: Both studies failed to demonstrate any benefit on overall survival (OS).",
        "\textcolor{blue}{2}: The aim of the analysis of these two trials was to identify subgroups of patients who may benefit from adjuvant CT."
        "\textcolor{blue}{3}: Individual patient data from two EORTC trials comparing doxorubicin-based CT to observation only in completely resected STS (large resection, R0/marginal resection, R1) were pooled.", ...
    ]' \\

    \textbf{Summary}: .\\
    \textbf{Citations}: .\\
    \midrule
    \textbf{Demonstrations} \\
    \textbf{Document}\\
    `[
        "\textcolor{blue}{0}: Giant cell tumor of bone (GCTB) is an aggressive primary osteolytic tumor.",
        "\textcolor{blue}{1}: GCTB often involves the epiphysis, usually causing substantial pain and functional disability.",
        "\textcolor{blue}{2}: Denosumab, a fully human monoclonal antibody against receptor activator of nuclear factor KB ligand (RANKL), is an effective treatment option for patients with advanced GCTB.",
        "\textcolor{blue}{3}: This analysis of data from an ongoing, open-label study describes denosumab's effects on pain and analgesic use in patients with GCTB. "
        "\textcolor{blue}{4}: Patients with unresectable disease (e.g. sacral or spinal GCTB, or multiple lesions including pulmonary metastases) were enrolled into Cohort 1 (N = 170), and patients with resectable disease whose planned surgery was associated with severe morbidity (e.g. joint resection, limb amputation, or hemipelvectomy) were enrolled into Cohort 2 (N = 101).", ...
    ]' \\
    \textbf{Summary}: \textcolor{myBlue}{The study aims to evaluate the effects of denosumab on pain and analgesic use in patients with giant cell tumor of bone (GCTB).}\\ 
    \textbf{Citations}: \textcolor{myBlue}{[3]} \\
    \\
    \hdashline
    \\
    \textbf{Document} \\ 
    `[
        "\textcolor{blue}{0}: Common adverse events associated with nivolumab included fatigue, pruritus, and nausea.",
        "\textcolor{blue}{1}: Drug-related adverse events of grade 3 or 4 occurred in 11.7\% of the patients treated with nivolumab and 17.6\% of those treated with dacarbazine."
        "\textcolor{blue}{2}: Nivolumab was associated with significant improvements in overall survival and progression-free survival, as compared with dacarbazine, among previously untreated patients who had metastatic melanoma without a BRAF mutation.",
        "\textcolor{blue}{3}: (Funded by Bristol-Myers Squibb; CheckMate 066 ClinicalTrials.gov number, NCT01721772.)."
    ]' \\
    \textbf{Summary}: \textcolor{gray}{Unknown.} \\
    \textbf{Citations}: \textcolor{gray}{Null.} \\
    \bottomrule
    \end{tabular}
    \captionof{table}{Instructions and demonstrations for generating summaries on aspect A (research aims). The \uwave{text} denotes placeholders to be replaced with aspect-specific descriptions.}
    \label{tab:llm_prompt}

\end{center}

\subsection{Instruction for summarizer $\mathcal{S}$ in \textsc{Track-Then-Sum}}
\begin{center}
    \small
    \begin{tabular}{p{\textwidth}}
    \toprule
    \textbf{Instructions} \\
    Summarize the \uwave{research aims or questions} of the study in one clear sentence that includes all key details from the input sentences without omitting important information.\\
    \\
    \textbf{Sentences} \\
    `[
        "The EORTC-STBSG coordinated two large trials of adjuvant chemotherapy (CT) in localized high-grade soft tissue sarcoma (STS).",
        "Both studies failed to demonstrate any benefit on overall survival (OS).",
        "The aim of the analysis of these two trials was to identify subgroups of patients who may benefit from adjuvant CT."
        "Individual patient data from two EORTC trials comparing doxorubicin-based CT to observation only in completely resected STS (large resection, R0/marginal resection, R1) were pooled."
    ]' \\
    \\
    \textbf{Summary:}\\
    \\
    \bottomrule
    \end{tabular}
    \captionof{table}{Instruction used to generate summaries for aspect A (research aims) in the summarization component of \textsc{Track-Then-Sum}. The \uwave{text} denotes placeholders to be replaced with aspect-specific descriptions.}
    \label{tab:prompt_tts}

\end{center}
\clearpage
\subsection{Instruction for summarizer $\mathcal{S}$ ($\oplus \ full \ context$) in \textsc{Track-Then-Sum}}
\begin{center}
    \begin{tabular}{p{\textwidth}}
    \toprule
    \textbf{Instructions} \\
    Summarize the \uwave{research aims or questions} of the study in one clear sentence that includes all key details from the input sentences without omitting important information. The summary must be based solely on the provided sentences. The full text is for reference only and must not be used to introduce any new information not present in the sentences. \\
    \\
    \textbf{Sentences}\\
    `[
        "The aim of the analysis of these two trials was to identify subgroups of patients who may benefit from adjuvant CT."
        "Individual patient data from two EORTC trials comparing doxorubicin-based CT to observation only in completely resected STS (large resection, R0/marginal resection, R1) were pooled."
    ]' \\
    \\
    \textbf{Full Context}\\
    `[
        "The EORTC-STBSG coordinated two large trials of adjuvant chemotherapy (CT) in localized high-grade soft tissue sarcoma (STS).",
        "Both studies failed to demonstrate any benefit on overall survival (OS).",
        "The aim of the analysis of these two trials was to identify subgroups of patients who may benefit from adjuvant CT."
        "Individual patient data from two EORTC trials comparing doxorubicin-based CT to observation only in completely resected STS (large resection, R0/marginal resection, R1) were pooled.", ...
    ]'\\
    \\
    \textbf{Summary:}\\
    \\
    \bottomrule
    \end{tabular}
    \captionof{table}{Instruction used to generate summaries for aspect A (research aims) in the summarization component of \textsc{Track-Then-Sum} ($\oplus \ f.$). The \uwave{text} denotes placeholders to be replaced with aspect-specific descriptions.}
    \label{tab:prompt_tts_f}
\end{center}

\vspace{0.5cm}

\subsection{Instruction for summarizer $\mathcal{S}$ in \textsc{Sum-Then-Track}}
\begin{center}
    \begin{tabular}{p{\textwidth}}
    \toprule
    \textbf{Instructions} \\
    Summarize the \uwave{research aims or questions} of the study in one clear sentence based on the given article. \\
    \\
    \textbf{Article}\\
    `[
        "The EORTC-STBSG coordinated two large trials of adjuvant chemotherapy (CT) in localized high-grade soft tissue sarcoma (STS).",
        "Both studies failed to demonstrate any benefit on overall survival (OS).",
        "The aim of the analysis of these two trials was to identify subgroups of patients who may benefit from adjuvant CT."
        "Individual patient data from two EORTC trials comparing doxorubicin-based CT to observation only in completely resected STS (large resection, R0/marginal resection, R1) were pooled.", ...
    ]'\\
    \\
    \textbf{Summary:} \\
    \\
    \bottomrule
    \end{tabular}
    \captionof{table}{Instruction used to generate summaries for aspect A (research aims) in the summarization component of \textsc{Sum-Then-Track}. The \uwave{text} denotes placeholders to be replaced with aspect-specific descriptions.}
    \label{tab:prompt_stt}
\end{center}

\vspace{0.5cm}

\subsection{Instruction for model $\mathcal{M}$ in \textsc{End-to-End}}
\begin{center}
    \begin{tabular}{p{\textwidth}}
    \toprule
    \textbf{Instructions} \\
    Given an article, summarize the \uwave{research aims or questions} of the study in one clear sentence and output the index of the cited sentences.\\
    \\
    \textbf{Sentences}\\
    `[
        "\textcolor{blue}{0}: The EORTC-STBSG coordinated two large trials of adjuvant chemotherapy (CT) in localized high-grade soft tissue sarcoma (STS).",
        "\textcolor{blue}{1}: Both studies failed to demonstrate any benefit on overall survival (OS).",
        "\textcolor{blue}{2}: The aim of the analysis of these two trials was to identify subgroups of patients who may benefit from adjuvant CT."
        "\textcolor{blue}{3}: Individual patient data from two EORTC trials comparing doxorubicin-based CT to observation only in completely resected STS (large resection, R0/marginal resection, R1) were pooled.", ...
    ]' \\
    \\
    \textbf{Summary:} \\
    \textbf{Citations:} \\
    \\
    \bottomrule
    \end{tabular}
    \captionof{table}{Instruction used to generate summaries for aspect A (research aims) in the \textsc{End-to-End}. The \uwave{text} denotes placeholders to be replaced with aspect-specific descriptions.}
    \label{tab:prompt_ete}
    \vspace{-15pt}
\end{center}

\end{document}